%% file: main.tex
\documentclass[runningheads]{llncs}

\usepackage{eccv}

\usepackage{eccvabbrv}

\usepackage{graphicx}
\usepackage{booktabs}

\usepackage[accsupp]{axessibility}  % Improves PDF readability for those with disabilities.

\usepackage{hyperref} 

\usepackage{colortbl}
\usepackage{orcidlink}
\usepackage{multirow}
\usepackage{pifont}
\newcommand{\cmark}{\ding{51}}%
\newcommand{\xmark}{\ding{55}}%
\usepackage{xcolor}
\definecolor{mycyan}{RGB}{100, 243, 235}
\usepackage{soul}
\usepackage{algorithm,multicol}
\usepackage[noend]{algpseudocode}
\usepackage{amsmath}
\usepackage{bm}
\usepackage{bbold}
\usepackage{makecell}

\begin{document}

\title{Long-Tail Temporal Action Segmentation with Group-wise Temporal Logit Adjustment} 

\titlerunning{Long-Tail TAS with Group-wise Temporal Logit Adjustment}

\author{Zhanzhong Pang\inst{1}\orcidlink{0009-0008-8320-1727} \and
Fadime Sener\inst{2}\orcidlink{0000-0001-5004-6005} \and
Shrinivas Ramasubramanian\inst{3}\orcidlink{0009-0002-0825-6775} \and
 Angela Yao\inst{1}\orcidlink{0000−0001−7418−6141}}

\authorrunning{Z. Pang, F. Sener, S. Ramasubramanian and A. Yao}

\institute{National University of Singapore \and
Meta Reality Labs \and
Carnegie Mellon University \\
\email{\email{\{pang, ayao\}@comp.nus.edu.sg}, famesener@meta.com, shrinivr@cs.cmu.edu}}

\maketitle
\input{sec/0_abstract}    
\input{sec/1_intro}
\input{sec/2_rel_works}
\input{sec/3_preliminaries}
\input{sec/4_method}
\input{sec/5_experiment}

\input{sec/6_conclusion}

\clearpage  % TODO REVIEW/FINAL: This \clearpage needs to be removed from both review and camera-ready versions.
\section*{Acknowledgements}
This research is supported by the Ministry of Education, Singapore, under the Academic Research Fund Tier 1 (FY2022).

\bibliographystyle{splncs04}
\bibliography{main}

\clearpage
\input{sec/7_suppl}

\end{document}

%% file: sec/0_abstract.tex
\begin{abstract}

% Temporal action segmentation assigns an action label to each frame in untrimmed procedural activity videos. Such 
Procedural activity videos often exhibit a long-tailed action distribution due to varying action frequencies and durations. However, state-of-the-art temporal action segmentation methods overlook the long tail and fail to recognize tail actions. Existing long-tail methods make class-independent assumptions and struggle to identify tail classes when applied to temporal segmentation frameworks. This work proposes a novel group-wise temporal logit adjustment~(G-TLA) framework that combines a group-wise softmax formulation while leveraging activity information  
and action ordering for logit adjustment. The proposed framework significantly  improves in segmenting tail actions without any performance loss on head actions. Source code is available\footnote{\url{https://github.com/pangzhan27/GTLA} }. 

\keywords{ Temporal action segmentation \and Procedural video understanding \and Long-tail recognition \and Logit adjustment}

\end{abstract}

%% file: sec/1_intro.tex
\section{Introduction} 
\label{sec:intro}
 
\begin{figure*}
\centering
\includegraphics[width=11.7cm, height=3.5cm]{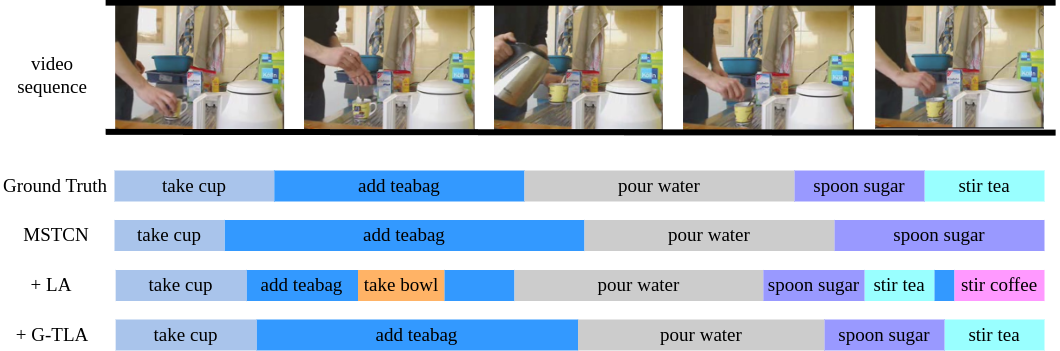}
\caption{ 
\emph{"Making tea"}, with temporal segments indicated by colored bars. The tail action~\textcolor{mycyan}{\emph{`stir tea'}} is recognized by Logit adjustment (LA) and our G-TLA but not by the MSTCN backbone.
%but  fails to recognize . While identify it, 
However, LA overlooks the action order and activity, resulting in activity-irrelevant false positives such as \textcolor{orange}{\emph{`take bowl'}} \& \textcolor{magenta}{\emph{`stir coffee'}}, and temporally illogical false positives like \textcolor{blue}{\emph{`add teabag'}} occurring after~\textcolor{mycyan}{ \emph{`stir tea'}}. 
} 
\label{fig:teaser}
\end{figure*}

\begin{figure*}[t] 
 \centering
 \begin{subfigure}{0.45\linewidth}
 \includegraphics[width=0.95\linewidth]{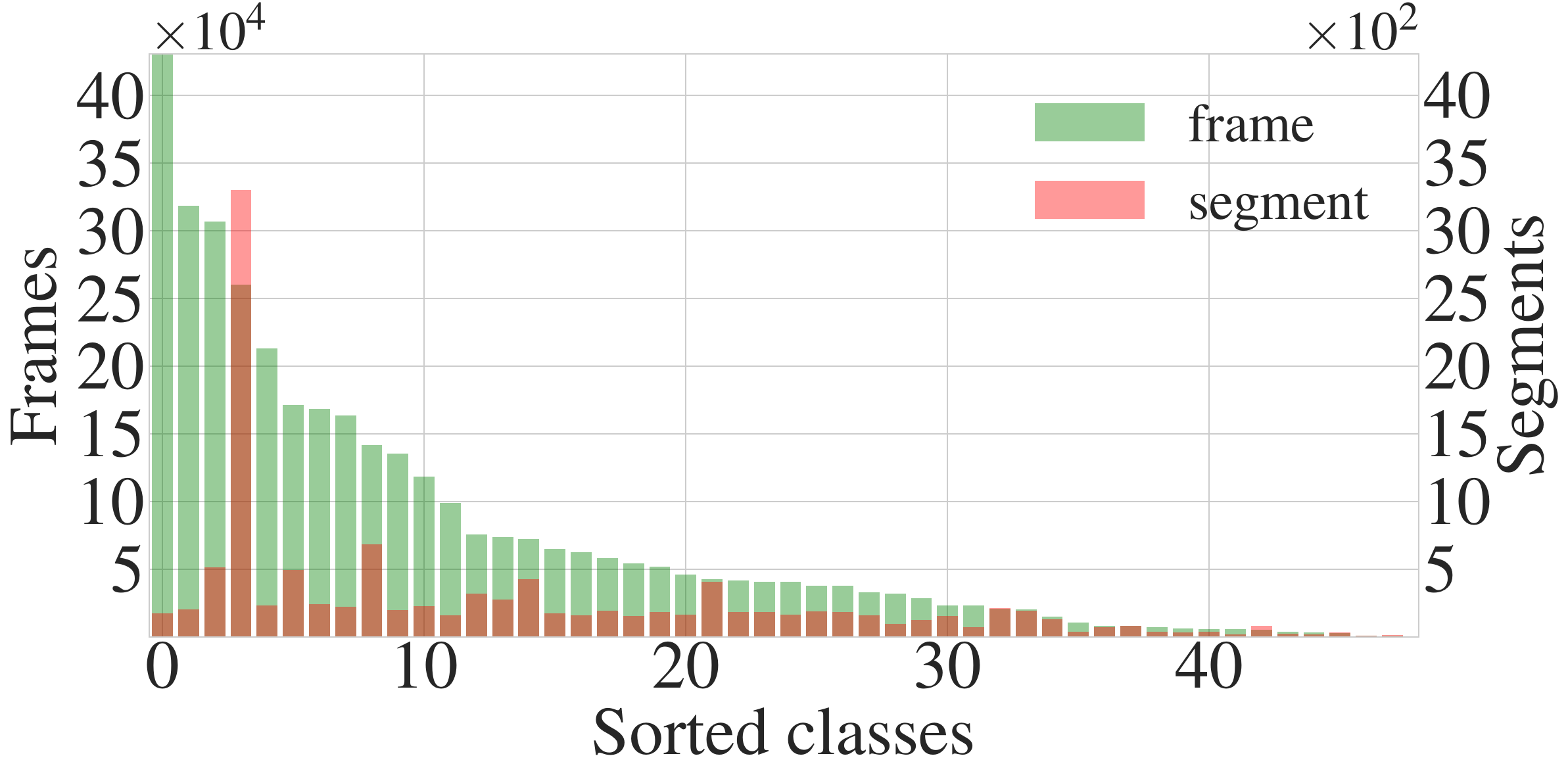}
 \caption{Breakfast.}
 \end{subfigure}
 \qquad
 \begin{subfigure}{0.45\linewidth}
 \includegraphics[width=0.95\linewidth]{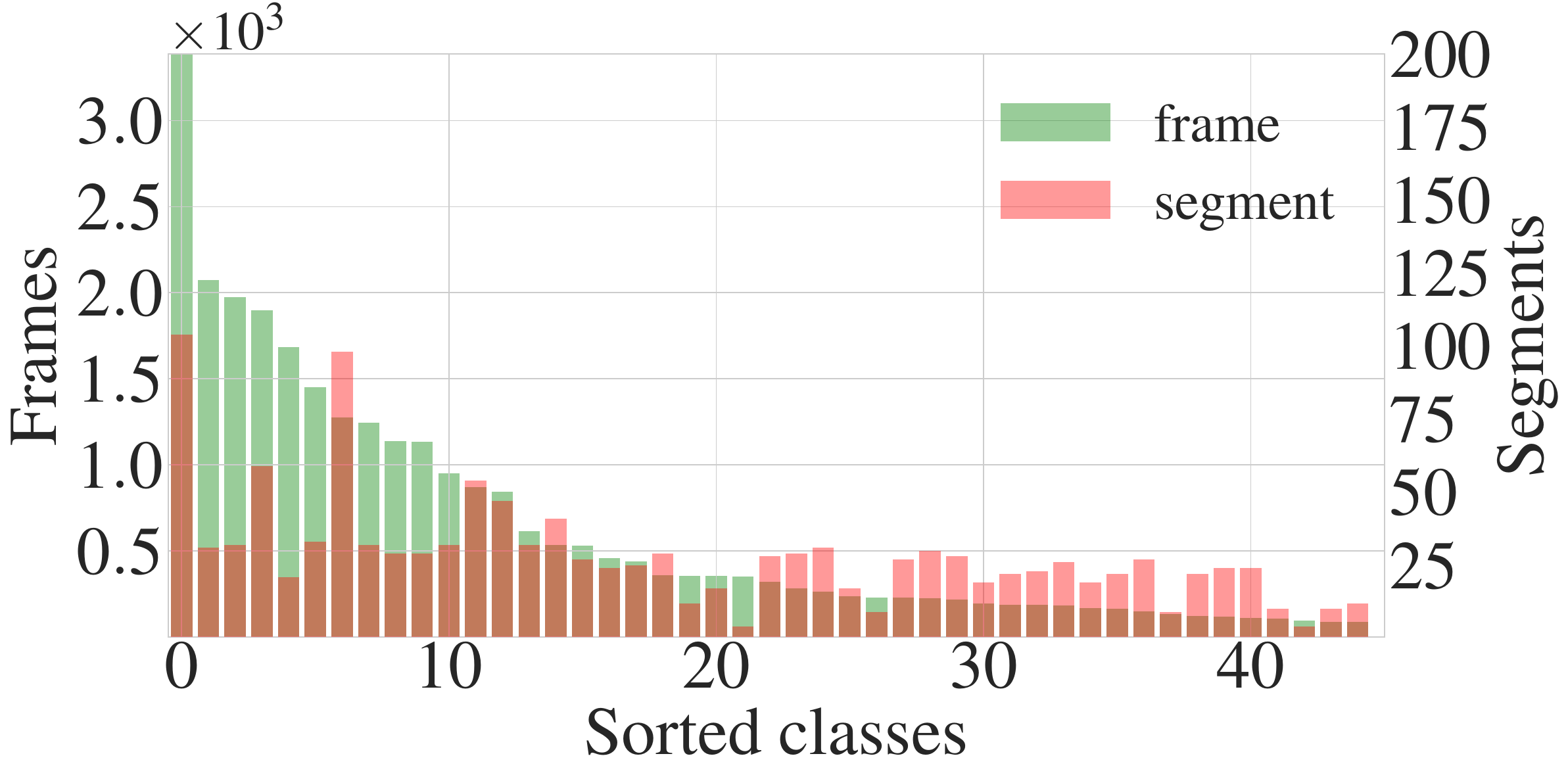}
 \caption{YouTube}
 \end{subfigure}
 \caption{Temporal action segmentation datasets exhibit a long-tail distribution of actions due to varying frequencies of actions and action durations.}
 \label{fig:data_dist}
\end{figure*}

Temporal action segmentation~\cite{stein2013,kuehne2014,li2020ms} partitions procedural activity sequences into multiple segments, each corresponding to a specific action class as depicted in \cref{fig:teaser}. There are two sources of tail in temporal action segmentation datasets. First, procedural videos often exhibit a long-tail distribution of action segments~\cite{ding2023temporal}, with infrequent actions forming the tail. For instance, in \cref{fig:teaser}, when \emph{"making tea"}, \emph{`add teabag'} is indispensable, while \emph{`spoon sugar'} and \emph{`stir tea'} are optional. The temporal action segmentation problem is mainly framed as a frame-wise classification task~\cite{farha2019ms,yi2021asformer,singhania2021coarse}, which leads to a second source of long-tail due to the varying durations of actions in procedural videos. For example, ~\emph{`pour water'} in \cref{fig:teaser} is longer and has more frames than \emph{`spoon sugar'}. The segment-~\&~frame-wise imbalances can be quite extreme, as depicted in \cref{fig:data_dist}.
 
The imbalance problem has been overlooked in temporal action segmentation literature~\cite{ding2023temporal,farha2019ms,yi2021asformer, singhania2021coarse, gao2021global2local}, leading to poor performance on tail classes\footnote{ASFormer~\cite{yi2021asformer}, DiffAct~\cite{liu2023diffusion} have zero accuracy on 5 and 4 of 48 classes, respectively.}. Our paper fills a significant gap by addressing the long-tail problem in temporal segmentation for the first time. Long-tail learning has mainly been investigated in  image~\cite{cui2019class,kang2019decoupling,menon2020long} and video classification~\cite{zhang2021videolt,perrett2023use}. Temporal segmentation differs, however, because actions are temporally correlated.  Yet conventional long-tail learning solutions such as re-sampling~\cite{han2005borderline,hu2020learning}, loss re-weighting~\cite{cui2019class,lin2017focal}, and logit adjustment (LA)~\cite{wang2021seesaw,menon2020long} make class independence assumptions. This compromises the learned temporal dependencies within the base models~\cite{li2020ms,yi2021asformer}; for example, LA incorrectly predicts \emph{`add teabag'} after \emph{`stir tea'} in ~\cref{fig:teaser}.  Moreover, for action segmentation, both frame- and segment-level performance is measured~\cite{farha2019ms}. Balancing head and tail classes, as well as frame- and segment-wise performance are two challenging trade-offs.  Conventional methods are ineffective in addressing the frame and segment trade-offs. For example, LA in ~\cref{fig:teaser} introduces irrelevant classes of \emph{`stir coffee'} and \emph{`take bowl'} from other activities, increasing false positives and %leading to 
over-segmentation. 

In procedural activities, some actions like \emph{`spoon sugar'} are shared across activities of \emph{"making tea"} and \emph{"making coffee"}; others are activity-specific, \eg action \emph{`stir tea'} occurs only in the activity of \emph{"making tea"}. Additionally, actions follow certain ordering: \emph{`stir tea'} always follows \emph{`pour water'}, even if other actions like \emph{`spoon sugar'} occur in between. Based on these observations, we propose a novel Group-wise Temporal Logit Adjustment (G-TLA) framework that encodes action order and activity information to address the long tail problem. Our framework consists of group-wise classification based on the activity label and temporal logit adjustment based on action ordering priors. Our method enhances tail recognition and mitigates over-segmentation by reducing false positives as shown in \cref{fig:fps}, including those from irrelevant activities and violating temporal dependencies. 
Our contributions can be summarized as: 
\begin{itemize}
\item the first method to address long-tail temporal action segmentation, along with our novel group-wise temporal logit adjustment (G-TLA) approach. 
\item leveraging class interdependencies to reduce false positives from long-tail learning, reducing over-segmentation. 
\item proposing new evaluation metrics to better reflect tail class performance, otherwise obscured by standard metrics. % that obscure the performance of tail classes. 
\item extensive evaluations on five datasets~\cite{kuehne2014language,alayrac2016unsupervised,fathi2011learning,stein2013combining,sener2022}, outperforming state-of-the-art backbones and standard long-tail learning approaches. 
\end{itemize}

%% file: sec/2_rel_works.tex
\section{Related works} 
\label{sec:related} 
 
\noindent\textbf{Temporal Action Segmentation.} Early approaches~\cite{bhattacharya2014,cheng2014temporal, fathi2011understanding} employed hand-crafted features to model frame-wise dependencies while using HMMs~\cite{kuehne2018hybrid} or RNNs~\cite{singh2016multi,perrett2017recurrent} to capture long-term dependencies. Subsequent methods leveraged pre-computed video features, such as I3D~\cite{carreira2017} and employed TCNs~\cite{lea2017temporal,li2020ms,lei2018temporal,farha2019ms,singhania2021coarse}, transformers~\cite{yi2021asformer} and diffusion models~\cite{liu2023diffusion} to learn frame- and action-wise relationships. For a comprehensive overview of temporal action segmentation, we refer to a recent survey~\cite{ding2023temporal}. To our knowledge, we are the first to address the long-tail problem in video sequences for temporal action segmentation~\cite{ding2023temporal}.

\noindent\textbf{Long-Tail Learning.} 
Long-tail methods can be broadly split into data and algorithm-level methods. Data-level methods either oversample the tail~\cite{buda2018systematic, shen2016relay, byrd2019effect} or undersample the head~\cite{he2009learning, drummond2003c4}. However, both approaches have drawbacks: oversampling can lead to tail overfitting, while undersampling is suboptimal for video datasets, as they often have limited samples. Additionally, applying existing data-level methods naively at the frame level is unsuitable for temporal action segmentation, as it requires preserving the sequence entirely as input. 

Algorithm-level methods include cost-sensitive learning, post-hoc adjustment, and ensembling techniques. Cost-sensitive learning balances loss functions by re-weighting different classes or samples~\cite{wang2017learning,huang2016learning,cui2019class,ren2018learning,lin2017focal}, or by adjusting the logits to enlarge margins for tail classes~\cite{menon2020long,cao2019learning,tan2020equalization,wang2021seesaw,zhao2022adaptive}. Post-hoc adjustment improves predictions for rare classes after training, \eg by normalizing the classification weights~\cite{kang2019decoupling,zhang2019balance,kim2020adjusting} or modifying the classification threshold~\cite{collell2016reviving,king2001logistic}. Ensembling combines multiple modules or experts, each specializing in different class distributions~\cite{zhou2020bbn,wang2020devil} or subsets~\cite{li2020overcoming,cai2021ace}. A key assumption in these methods is class independence. However, in action segmentation, videos can be assumed to be \emph{i.i.d.}, while the actions within each video exhibit dependencies. Our work is the first to consider these dependencies in designing a long-tailed solution.

%% file: sec/3_preliminaries.tex
\section{Preliminaries}
\label{sec:pre}
 
\subsection{Temporal Action Segmentation}
Temporal action segmentation maps frame-wise representations, $\mathcal{X}$, to action labels, $\mathcal{Y}=[L]=\{1,2, \dots, L\}$. Consider a video sample, $\{X, Y\}$: $X = \{x_t\} \in \mathbb{R}^{D \times T}, Y = \{y_t\} \in [L]^{T}$, where $D$ is the feature dimension, $T$ is the number of frames in a video, $t$ is the frame index, $L$ is the number of classes. In line with previous works~\cite{farha2019ms, yi2021asformer, singhania2021coarse}, we use pre-computed video representations~\cite{carreira2017} as $X$. 
 
Consider a classifier, denoted as $f$, which takes the entire sequence, $X$, as input to learn temporal correlations among frames and outputs a sequence of action labels, $Y$. Classifier $f$ is a neural network such as MSTCN~\cite{farha2019ms} or AsFormer~\cite{yi2021asformer}. Classifier $f: \mathcal{X} \rightarrow \mathcal{Y}$ is trained by minimizing a frame-wise cross-entropy loss:
\begin{align}
 \mathcal{L}_{cls} = \frac{1}{T} \sum_t -\log \hat{p}(y_t),
\label{eq: cls_lss}
\end{align}
where $\hat{p}(y_t)$ is the estimated probability for the ground truth label $y_t$ of frame $x_t$. A smoothing loss is commonly applied with threshold $\delta$ to encourage smooth transitions between frames. Following prior works~\cite{farha2019ms}, we set $\delta$ to 4. 
\begin{align}
 \mathcal{L}_{sm} &= \frac{1}{TL} \sum_{t, c} \hat{\triangle}^2_{t,c}, \ \ \hat{\triangle}_{t,c} = \begin{cases} \triangle_{t,c}, & \triangle_{t,c} \le \delta \\ \ \ \ \delta \quad , & \text{otherwise} \end{cases} \nonumber \\
 \triangle_{t,c} & = |\log \hat{p}(c_t) - \log \hat{p}(c_{t-1})|, \ c \in [1, \cdots, L],
\label{eq: sm_lss}
\end{align}
where $\hat{p}(c_t)$ is the estimated probability of class $c$ at time $t$.

\subsection{Logit Adjustment}
Given a classification problem with data-label pairs $\{x,y\}$, the standard training objective is to learn a model $f$ that minimizes the expected error $ \textbf{E}_{x,y} \ \text{Err}(f(x),y)$. When the label distribution is highly skewed, the learning process minimizes a skewed error, prioritizing classes with more samples. In such scenarios, a balanced error $ \textbf{E}_{x \mid y} \ \text{Err}(f(x),y)$ that averages the per-class loss~\cite{brodersen2010balanced, menon2013statistical} is more suitable. Given the prior $p(c)$ and the posterior $p(c\!\mid\!x)$ of class $c$, the optimal classifier for minimizing the balanced error~\cite{collell2016reviving} takes the form
\begin{equation}
 \arg\max_{c} \frac{p(c \mid x)} {p(c)} \approx \arg\max_{c} s_c(x) -\log p(c)
\label{eq: la}
\end{equation}
where a neural network typically estimates $p(c\!\mid\!x)$ as a logit $s_c(x)$. For simplicity, we omit the logit's dependency on the neural network's parameters. 

\cref{eq: la} is the vanilla logit adjustment~\cite{menon2020long}, which %differs from standard learning by 
incorporating the prior during inference. Logit adjustment can be further incorporated into training by enforcing a class prior offset while learning the logits~\cite{menon2020long}.
\begin{align}
 \tilde{s}_c(x) &= s_c(x) + \tau \log p(c),  \quad \tilde{p}(c|x) = \text{softmax}(\tilde{s}_c(x))\label{eq: la_train0}
\end{align}
where $s_c(x)$ is the output logit, $\tilde{s}_c(x)$ is the adjusted logit, $\tilde{p}(c|x)$ is the predicted probability of class $c$ after adjusting logit and used in the cross-entropy loss(we use $p$ to represent the prior, $\hat{p}$ for predicted probability, and $\tilde{p}$ for probability after adjusting logit). $\tau$ controls the trade-off between minimizing balanced and skewed errors. Introducing the adjustment adds a per-class margin into the softmax, shifting the decision boundary away from tail classes as in \cref{eq: la_boundary}.
\begin{equation}
 \mathcal{L}(y, f(x)) = - \log \frac{e^{s_y(x) + \tau \log p(y)}}{\sum_c e^{s_c(x) + \tau \log p(c)}} = \log \left[ 1 + \sum_{c\neq y} e^{s_c(x) - s_y(x) + \tau \frac{\log p(c)}{\log p(y)}} \right]
\label{eq: la_boundary}
\end{equation}
 

%% file: sec/4_method.tex
\section{Methodology}
\label{sec:method}
Our method leverages action order and activity information to address the long-tail problem in action segmentation. We begin by introducing the action inter-dependencies in \cref{sec:actioninter} and describe how to solve such dependencies through group-wise classification in \cref{sec: group} and temporal logic adjustment in \cref{sec: tla}. Details of training and inference are presented in \cref{sec: train} and \cref{sec: test}.

\subsection{Action Inter-Dependencies}
\label{sec:actioninter} 
The independent assumption of prior~$p(c)$ in~\cref{eq: la_train0} does not hold for actions in videos.
In procedural activities, actions within a sequence interact. Accurately estimating the prior of action $c$ conditioned on sequence $Y$, $p(c|Y)$ is challenging due to varying action orders and limited training samples. Notably, the class distribution~$p(c)$ is activity,~$a$, dependent; naively using $p(c)$ leads to activity-irrelevant false positives as in~\cref{fig:teaser}. Therefore, we propose a relaxed solution for the temporal prior as $p(c|a)$  to condition on activity label $a$. 
As the video sequences have a loose ordering of actions, we assume that action classes are conditionally independent given the activity label. We further exploit useful action ordering in \cref{sec: tla}. 

Given an input sequence, $X$, and its activity label, $a$, frame predictions are independent while frame representations interact~\cite{farha2019ms, yi2021asformer}, \ie frame prediction $\hat{y}_t$ is based on the sequence representation $X$ instead of a single frame representation $x_t$.  Then, the logit adjustment for frame $t$, according to \cref{eq: la_train0}, is:

\begin{equation}
 s_{c,t}(X) + \tau \log p(c \mid a),
\label{eq: la_tas_s}
\end{equation}
where $s_{c,t}(X)$ is the logit of class $c$ at time $t$. 

\subsection{Group-wise Classification}
\label{sec: group}
To leverage the conditional prior $p(c|a)$, we propose a hierarchical classifier
by categorizing sequences into disjoint groups and assigning actions within each sequence to the corresponding group.  This grouping strategy enables group-wise action classification and logit adjustment based on the group's prior. % conditioned on the group. 
Additionally, grouping ensures no incompatible actions in each activity group. For example, the action \emph{`stir coffee'} does not occur in activities like \emph{"making tea"}. The incompatibility causes zero conditional probability $p(c|a) = 0$,  posing a numerical problem for logit adjustment. Adding a small value to $p(c|a)$ does not solve the numerical problem. Instead, it converts activity-incompatible classes to tail classes. These classes are subsequently overemphasized and result in more activity-irrelevant false positives.
 
\begin{figure}[t]
 \centering
 \includegraphics[width=0.8\linewidth, height = 6.0cm]{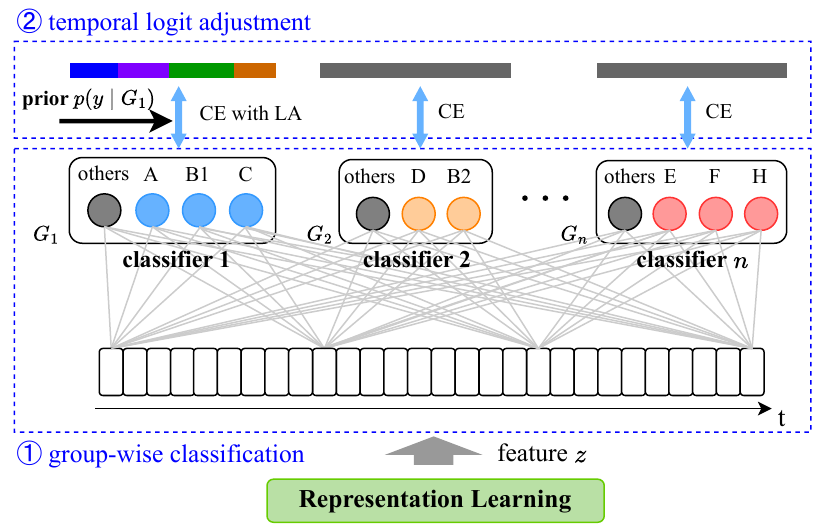}
 \caption{Our group-wise temporal logit adjustment framework consists of group-wise classification and temporal logit adjustment within the respective group. The temporal logit adjustment is only applied to the target group($G_1$ in this illustration).}
 \label{fig:framework}
\end{figure}
 
We denote the group set as $\mathbf{G}$. Each group may contain activities with shared actions. While a straightforward approach is to group based on the activity label, we also explore sequence clustering without activity labels based on the KL-divergence of action frequency distributions(see Supplementary for details). For each group, we introduce an auxiliary class, \emph{`others'}, corresponding to actions that do not belong to the current group. Classes shared across different groups are treated as separate classes. For example, \emph{`spoon sugar'} occurs in both \emph{"making coffee"} and \emph{"making tea"} which are different groups ${G_1}$ and $G_2$, then \emph{`spoon sugar'} is treated as different classes $y^{(1)}$ and $y^{(2)}$. Non-shared classes, like \emph{`stir tea'}, are labeled as \emph{`stir tea'} in $G_2$, and \emph{`others'} in the remaining groups.
 
Our group-wise classification strategy is illustrated in \cref{fig:framework}. The final feature layer $z_t$ at time $t$ is fed into $n$ classifiers, where $n$ is the number of groups in $\mathbf{G}$. These classifiers generate predictions concurrently based on the last feature layer. For the $i^\text{-th}$ group, the predicted logit of class $c$ at time $t$ is computed as 
\begin{equation}
 s_{c,t}^{(i)}(X)= \sum_j z_t[j] \cdot W_{j,c}^{(i)} + b^{(i)}
\label{eq: group_head}
\end{equation}
where $W^{(i)}$ and $b^{(i)}$ are the classifier weights and bias for group $i$.

The overall loss is the sum of two cross-entropies: one for the actions in the target group and one for the auxiliary class for non-target groups:

\begin{equation}
 \mathcal{L}_{G} = \alpha_k \frac{1}{T} \sum_t -\log \hat{p}(y_t^{(k)}) + \eta \sum_{i \neq k}^n \frac{1}{T} \sum_t -\log \hat{p}(o_t^{(i)})
\label{eq: group_loss}
\end{equation} 
where $k$ is the current sequence's group, $\hat{p}(y_t^{(k)})$ represents the predicted probability of the truth label $y_t$ in the target group $k$, $\hat{p}(o_t^{(i)})$ is the predicted probability of class \emph{`others'} in the non-targeted group. $\eta$ is a hyperparameter that balances the losses between target and non-target groups and $\alpha_k$ represents a pre-computed reweighting factor designed to balance the bias arising from group size, \ie the number of sequences in each group. From a gradient analysis perspective~\cite{wang2021seesaw}, a small $\eta$ helps reduce the suppression of tail classes by down-weighting the gradients from negative samples of the class \emph{others}. However, if $\eta$ is too small, it may hinder group identification during inference.
 
Group-wise classification offers several advantages. Separating semantically similar actions into different groups mitigates confusion, \eg distinguishing between \emph{`take plate'} and \emph{`take cup'} becomes easier, despite sharing a verb. Additionally, it reduces false positives from irrelevant classes, ensuring that only group-compatible actions are predicted, preventing implausible predictions.

\subsection{Temporal Logit Adjustment}
\label{sec: tla}

Our group-wise framework simplifies frame-wise classification. 
With group information we can substitutes $p(c|a)$ in \cref{eq: la_tas_s} with $p(c|G_k)$.
However, there is still valuable sequential ordering information. For example, the action \emph{`stir tea'} always follows \emph{`pour water'}, even if other actions \eg \emph{`spoon sugar'} occur in between. This ordering information can also be incorporated into the logit adjustment. Therefore, we adopt a fine-grained approach by tailoring the adjustment to each class $c$ and frame $t$ within the target group $G_k$, as follows:
\begin{equation}
 s_{c, t}^{(k)}(X) + \tau \mathcal{T}_{c,t}^{(k)}(X) \log p(c \mid G_k),
\label{eq: la_tas}
\end{equation}
where the temporal factor $\mathcal{T}^{(k)}_{c,t}(X)$ refines the logit adjustment based on the temporal bounds for class $c$ in the target group $G_k$.

\textbf{Temporal Factor.} 
For an action $c$, we define $S_{bf}[c]$ as the set of actions that must precede $c$, and $S_{af}[c]$ that must follow $c$. These two sets are exclusive and can be derived from the training data. Given a sequence $X$ from group $G_k$, we can utilize the two sets to determine the temporal bounds $[t_1(c, X), t_2(c, X)]$ within which action $c$ can occur, where $t_1(c, X)$ and $t_2(c, X)$  are determined by the latest and earliest occurrence time of classes in $S_{bf}[C]$ and $S_{af}[C]$, respectively.
\begin{equation}
\resizebox{1.04\hsize}{!}{$
 t_1(c, X) = \begin{cases} \max_t y_t \in S_{bf}[c], \ \text{if} \ S_{bf}[c] \neq \emptyset\\ 0, \ \text{otherwise} \end{cases}, 
 t_2(c, X) = \begin{cases} \min_t y_t \in S_{af}[c], \ \text{if} \ S_{af}[c] \neq \emptyset \\ T , \ \text{otherwise}, 
\end{cases}$}
\label{eq: time}
\end{equation}
where $y_t$ is the ground truth label of $X$ at time $t$.

\begin{figure}[t]
 \centering
 \includegraphics[width=0.83\linewidth, height = 1.68cm]{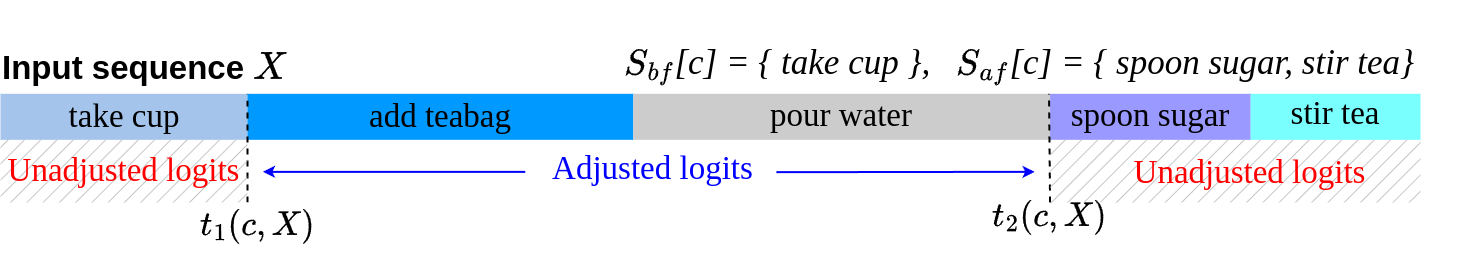}
 \caption{Illustration of temporal logit adjustment for class $c=$  \emph{`add teabag'}. The adjustment only occurs within the temporal bounds.}
 \label{fig:TLA}
\end{figure}

Temporal factor adjusts the logit for class $c$ in the target group $G_k$ based on these temporal bounds. Logits within the temporal bounds are adjusted normally, while those outside the bounds are left unadjusted to prevent false positives that violate the temporal prior  as in \cref{fig:TLA}. This can be formalized as:
\begin{equation}
 \mathcal{T}^{(k)}_{c,t}(X) = \begin{cases} 1 ,& \text{if} \ t \in [t_1(c, X), t_2(c, X)] \\ \frac{\log p(y_t | G_k)}{\log p(c | G_k)} ,& \text{otherwise}, \end{cases} 
\label{eq: temporal}
\end{equation}
where \emph{otherwise}, we use the ratio of the prior for label $y_t$ and class $c$ to ensure consistent adjustment with the target label, $p(y_t \mid G_k)$. This prevents misclassification due to adjustment between $y_t$ and $c$, avoiding violating temporal prior. 
 
\subsection{Overall loss}
\label{sec: train}
With group-wise classification and temporal logit adjustment, we can reframe the classification loss~\cref{eq: group_loss} as
\begin{align}
 \mathcal{L}_{GTLA} = \alpha_k \frac{1}{T} \sum_t -\log \tilde{p}(y_t^{(k)}) + \eta \sum_{i \neq k}^n \frac{1}{T} \sum_t -\log \hat{p}(o_t^{(i)}),
\label{eq: glta_loss}
\end{align}
where predicted probabilities $\tilde{p}$ for the target group $k$ is based on adjusted logits in \cref{eq: la_tas}.  For non-target groups $i$, logit adjustment is not performed; instead, the naive logit $s^{(i)}_{o, t}(X)$ of class \emph{`others'} is used to calculate the probability.
 
The training loss $\mathcal{L}$ combines group-wise logit adjustment loss $\mathcal{L}_{GTLA}$ and smooth loss $\mathcal{L}_{sm}$,  $\mathcal{L} = \mathcal{L}_{GTLA} + \lambda \mathcal{L}_{sm}$, with $\lambda$ as the trade-off hyper-parameter. %where $\lambda$ is the trade-off hyper-parameter balancing the terms. 

\subsection{Inference} 
\label{sec: test}
The group label of a sequence is unknown during inference. We use the prediction for \emph{`others'} to identify the group. Specifically, the group with the lowest probability for \emph{`others'} across all frames is selected as the predicted group $\hat{k}$(\cref{eq: group_infer}).
The final result is the prediction (excluding class \emph{`others'}) from the classifier corresponding to the identified group. During inference, temporal logit adjustment is not used, \ie inference applies argmax on predicted probabilities $\hat{p}$ instead of adjusted probability $\tilde{p}$ through temporal logit adjustment. 
\begin{equation}
 \hat{k} = \arg\min_i \frac{1}{T} \sum_t \hat{p}(o^{(i)}_{t}), \quad \hat{y}_t = \arg\max_{c_t^{(\hat{k})} \neq \text{others}} \hat{p}(c_t^{(\hat{k})}), 
\label{eq: group_infer}
\end{equation} 
where $c_t^{(\hat{k})}$ represent class $c$ in group $\hat{k}$ at time $t$.

%% file: sec/5_experiment.tex
\section{Experiments}
\label{sec:experiments}

\subsection{Dataset, Implementation, and Evaluation}
\noindent \textbf{Dataset.} We evaluate our method on five datasets: Breakfast Actions~\cite{kuehne2014language} YouTube Instructional Videos~\cite{alayrac2016unsupervised}, Assembly101~\cite{sener2022}, GTEA~\cite{fathi2011learning} and 50Salads~\cite{stein2013combining}. We follow the same settings and splits in~\cite{farha2019ms,sener2022}. We split classes in these datasets into Head and Tail according to their frequency, with distributions for Breakfast and YouTube shown in \cref{fig:data_dist} and others in the Supplementary.
\begin{table}[htb]
\centering
\caption{Class split and imbalance of used datasets. The imbalance ratio is the number of samples in the most frequent class divided by the number in the least frequent class.}
\begin{tabular}{c|c|c|c|c||c}
    \toprule
    \multirow{2}{*}{\textbf{Dataset}}  & \multicolumn{2}{c|}{\textbf{Head group}} & \multicolumn{2}{c||}{\textbf{Tail group}} & \multirow{2}{*}{\textbf{Imbalance ratio}} \\
    \cline{2-5} & $\#$actions & $\#$frames  & $\#$actions & $\#$frames \\
    \midrule
    Breakfast & 20 & $\ge$ 5$\times 10^4$ & 28 & $\le$ 5$\times 10^4$ & 639 \\
    YouTube & 14 & $\ge$ 500 & 32 & $\le$ 500 & 558\\
    Assembly101 & 31 & $\ge$ 1.8 $\times 10^5$ &  171 & $\le$ 1.8 $\times 10^5$  & 2604 \\ 
    50salads & 6 & $\ge$ 4$\times 10^4$ & 13 & $\le$ 4$\times 10^4$ & 6\\
    GTEA & 5 & $\ge$ 2000 & 6 & $\le$ 2000 & 24\\
    \bottomrule
  \end{tabular}
  \label{tab:cls_split_supp}
\end{table}

\noindent \textbf{Implementation Details.} We use a temporal convolution model, MSTCN~\cite{farha2019ms}, and a transformer model, ASFormer~\cite{yi2021asformer} as backbones. We retrain these models using the same protocols, settings, and pre-extracted I3D features as the original papers; see the Supplementary for details and hyper-parameters. During training, activity labels or clustering results serve as group labels. During testing, the group label is inferred from predictions of the class \emph{`others'} in that group.

\begin{table}[htb]
\caption{Comparison on YouTube with harmonic mean on head and tail classes over three runs. (Global metrics in grey are for reference. See Suppl. for standard deviation).}
\centering 
\setlength{\tabcolsep}{2mm}{
\resizebox{1.01\columnwidth}{!}{
\begin{tabular}{l|l|lll|lll|ll}
\hline
\multirow{2}{*}{\textbf{Model}} & \multirow{2}{*}{\textbf{Type}} & \multicolumn{3}{c|}{\textbf{Frame acc}} & \multicolumn{3}{c|}{\textbf{Segment F1@25}} & \multicolumn{2}{c}{\textcolor{gray}{\textbf{Global}}} \\
\cline{3-10} & & \textbf{Head} & \textbf{Tail} & \textbf{Hmean} & \textbf{Head} & \textbf{Tail} & \textbf{Hmean}  & \textcolor{gray}{\textbf{Acc}} & \textcolor{gray}{\textbf{F1@25}} \\
 \hline
 \textbf{AsFormer} & -                              & 53.1 & 17.2 & 26.0 & 47.6 & 20.2 & 28.4 & \textcolor{gray}{69.8} & \textcolor{gray}{45.6}\\
 + CB~\cite{cui2019class} & reweight                & \textcolor{red} {\ -2.2} & \textcolor{blue} {+2.9} & \textcolor{blue} {+2.8} & \textcolor{red} {\ -0.9} & \textcolor{blue} {+1.0} & \textcolor{blue} {+0.8} & \textcolor{gray} {\ -0.2} & \textcolor{gray} {\ -0.2}\\
 + Focal~\cite{lin2017focal} & reweight             & \textcolor{red} {\ -2.4} & \textcolor{blue} {+0.8} & \textcolor{blue} {+0.1} & \textcolor{red} {\ -2.2} & \textcolor{blue} {+2.0} & \textcolor{blue} {+1.4} & \textcolor{gray} {\ -0.1} & \textbf{\textcolor{gray} {+1.0}}\\
 + BAGS~\cite{li2020overcoming} & ensemble          & \textcolor{red} {\ -1.2} & \textcolor{blue} {+3.2} & \textcolor{blue} {+3.3} & \textcolor{red} {\ -0.8} & \textcolor{blue} {+1.8} & \textcolor{blue} {+1.6} & \textcolor{gray} {\ -0.5} & \textcolor{gray} {\ -0.5}\\
 + $\tau$-norm~\cite{kang2019decoupling} & post-hoc & \textcolor{blue} {+1.2} & \textcolor{blue} {+1.3} & \textcolor{blue} {+1.6} & \textcolor{red} {\ -1.0} & \textcolor{blue} {+0.3} & \textcolor{blue} {+0.1} & \textcolor{gray} {\ -0.8} & \textcolor{gray} {\ -1.3}\\
 + LA~\cite{menon2020long} & logit adj.             & \textcolor{blue} {+0.8} & \textcolor{blue} {+4.9} & \textcolor{blue} {+5.3} & \textcolor{red} {\ -0.7} & \textcolor{blue} {+2.3} & \textcolor{blue} {+2.1} & \textcolor{gray} {\ -1.9} & \textcolor{gray} {\ -0.5}\\
 + LDAM~\cite{cao2019learning} & logit adj.         & \textcolor{blue} {+1.2} & \textcolor{blue} {+3.6} & \textcolor{blue} {+3.4} & \textcolor{red} {\ -1.7} & \textcolor{blue} {+1.4} & \textcolor{blue} {+0.9} & \textcolor{gray} {\ -0.8} & \textcolor{gray} {\ -0.8}\\
 + Seesaw~\cite{wang2021seesaw} & logit adj.        & \textcolor{red} {\ -0.6} & \textcolor{blue} {+1.9} & \textcolor{blue} {+2.0} & \textcolor{red} {\ -1.9} & \textcolor{blue} {+1.3} & \textcolor{blue} {+0.8} & \textcolor{gray} {\ -0.6} & \textcolor{gray} { +0.1}\\
 + G-TLA(ours) & logit adj. & \textbf{\textcolor{blue} {+2.3}} & \textbf{\textcolor{blue} {+6.8}} & \textbf{\textcolor{blue} {+7.5}} & \textbf{\textcolor{red} {\ -0.3}} & \textbf{\textcolor{blue} {+5.1}} & \textbf{\textcolor{blue} {+4.6}} & \textbf{\textcolor{gray} {+0.1}} & \textcolor{gray} {+0.6}\\
 \hline
 \textbf{MSTCN} & -                                 & 46.0 & 15.5 & 23.2 & 39.0 & 16.8 & 23.5 & \textcolor{gray} {68.0} & \textcolor{gray} {39.1}\\
 + CB~\cite{cui2019class} & reweight                & \textcolor{red} {\ -0.2} & \textcolor{blue} {+2.5} & \textcolor{blue} {+2.7} & \textcolor{blue} {+1.2} & \textcolor{red} {\ -0.4} & \textcolor{red} {\ -0.2} & \textcolor{gray} {\ -1.7} & \textcolor{gray} {\ -0.4}\\
 + Focal~\cite{lin2017focal} & reweight             & \textcolor{blue} {+1.7}  & \textcolor{blue} {+1.5} & \textcolor{blue} {+1.8} & \textcolor{blue} {+1.9} & \textcolor{blue} {+1.2} & \textcolor{blue} {+1.9} & \textcolor{gray} {\ -0.5} & \textcolor{gray} {+0.9}\\
 + BAGS~\cite{li2020overcoming} & ensemble          & \textcolor{red} {\ -0.4} & \textcolor{blue} {+2.1} & \textcolor{blue} {+2.1} & \textcolor{blue} {+2.5} & \textcolor{blue} {+0.5} & \textcolor{blue} {+0.9} & \textcolor{gray} {\ -0.8} & \textcolor{gray} {+1.0}\\
 + $\tau$-norm~\cite{kang2019decoupling} & post-hoc & \textcolor{blue} {+0.5}  & \textcolor{blue} {+1.1} & \textcolor{blue} {+1.3} & \ 0.0                   & \textcolor{red} {\ -0.6} & \textcolor{red} {\ -0.6} & \textcolor{gray} {\ -0.7} & \textcolor{gray} {\ -1.1}\\
 + LA~\cite{menon2020long} & logit adj.             & \ 0.0                    & \textcolor{blue} {+2.1} & \textcolor{blue} {+2.3} & \textcolor{blue} {+0.4} & \textcolor{red} {\ -0.8} & \textcolor{red} {\ -0.7} & \textcolor{gray} {\ -1.0} & \textcolor{gray} {\ -0.3}\\
 + LDAM~\cite{cao2019learning} & logit adj.         & \textcolor{red} {\ -2.3} & \textcolor{blue} {+0.3} & \ 0.0                   & \textcolor{red} {\ -2.1} & \textcolor{blue} {+0.6} & \textcolor{blue} {+0.2} & \textcolor{gray} {\ -0.4} & \textcolor{gray} {+0.1}\\
 + Seesaw~\cite{wang2021seesaw} & logit adj.        & \textcolor{blue} {+1.8}  & \textcolor{blue} {+1.4} & \textcolor{blue} {+1.8} & \textcolor{blue} {+1.8} & \textcolor{blue} {+0.3} & \textcolor{blue} {+0.6} & \textbf{\textcolor{gray} {\ -0.1}} & \textcolor{gray} {+0.6}\\
 + G-TLA(ours) & logit adj. & \textbf{\textcolor{blue} {+2.7}} & \textbf{\textcolor{blue} {+6.3}} & \textbf{\textcolor{blue} {+6.8}}& \textbf{\textcolor{blue} {+2.7}} & \textbf{\textcolor{blue} {+3.3}} & \textbf{\textcolor{blue} {+3.6}} & \textcolor{gray} {\ -0.4} & \textbf{\textcolor{gray} {+1.1}}\\ 
\hline
\end{tabular}
}} 
\label{tab:summary2}
\end{table}

\begin{table}[htb]
\caption{Comparison on Breakfast with harmonic mean on head and tail classes over three runs. (Global metrics in grey are for reference. See Suppl. for standard deviation).}
\centering 
\setlength{\tabcolsep}{2mm}{
\resizebox{1.01\columnwidth}{!}{
\begin{tabular}{l|l|lll|lll|ll}
\hline
\multirow{2}{*}{\textbf{Model}} & \multirow{2}{*}{\textbf{Type}} & \multicolumn{3}{c|}{\textbf{Frame acc}} & \multicolumn{3}{c|}{\textbf{Segment F1@25}} & \multicolumn{2}{c}{\textcolor{gray}{\textbf{Global}}} \\
\cline{3-10} & & \textbf{Head} & \textbf{Tail} & \textbf{Hmean} & \textbf{Head} & \textbf{Tail} & \textbf{Hmean}  & \textcolor{gray}{\textbf{Acc}} & \textcolor{gray}{\textbf{F1@25}} \\
 \hline
 \textbf{AsFormer}        & -                       & 69.7 & 39.8 & 50.7 & 69.9 & 43.9 & 53.9 & \textcolor{gray}{72.4} & \textcolor{gray}{69.9} \\
 + CB~\cite{cui2019class} & reweight                & \textcolor{blue} {+0.5} & \textcolor{blue} {+1.0} & \textcolor{blue} {+0.9} & \textcolor{blue} {+1.3} & \textcolor{blue} {+0.8} & \textcolor{blue} {+1.0}  & \textcolor{gray} {\ -0.5}  & \textcolor{gray} {\ -0.2}\\
 + Focal~\cite{lin2017focal} & reweight             & \textcolor{blue} {+0.2} & \textcolor{red} {\ -0.7} & \textcolor{red} {\ -0.5} & \textcolor{blue} {+1.4} & \textcolor{red} {\ -0.2} & \textcolor{blue} {+1.2} & \textcolor{gray} {\ -0.1}  & \textcolor{gray} {+0.5}\\
 + BAGS~\cite{li2020overcoming} & ensemble          & \textcolor{red} {\ -0.2} & \textcolor{blue} {+0.8} & \textcolor{blue} {+0.5} & \textcolor{blue} {+1.6} & \textcolor{blue} {+0.5} & \textcolor{blue} {+1.0} & \textcolor{gray} {\ -0.6}  & \textcolor{gray} {\ -1.0}\\
 + $\tau$-norm~\cite{kang2019decoupling} & post-hoc & \textcolor{red} {\ -0.1} & \textcolor{blue} {+1.3} &\textcolor{blue} {+0.8} &  \ 0.0 & \textcolor{blue} {+0.4} & \textcolor{blue} {+0.3} & \textcolor{gray} {\ -0.2}  & \textcolor{gray} {\ -0.8}\\
 + LA~\cite{menon2020long} & logit adj.             & \textcolor{blue} {+0.4} & \textcolor{blue} {+0.6}  & \textcolor{blue} {+0.6} & \textcolor{blue} {+0.3} & \textcolor{blue} {+0.8} & \textcolor{blue} {+1.0} & \textcolor{gray} {+0.1}  & \textcolor{gray} {\ -0.2}\\
 + LDAM~\cite{cao2019learning} & logit adj.         & \textcolor{blue} {+0.4}  & \textcolor{blue} {+1.2} & \textcolor{blue} {+1.0} & \textcolor{blue} {+0.3} & \textcolor{blue} {+1.1} & \textcolor{blue} {+1.2} & \textbf{\textcolor{gray} {+0.2}} & \textcolor{gray} {+0.7} \\
 + Seesaw~\cite{wang2021seesaw} & logit adj.        & \textcolor{blue} {+0.4}  & \textcolor{blue} {+1.3} & \textcolor{blue} {+1.1} & \textcolor{blue} {+0.5} & \textcolor{blue} {+1.5} & \textcolor{blue} {+1.6} & \textcolor{gray} {+0.1} & \textcolor{gray} {+0.2} \\
 + G-TLA(ours) & logit adj.                         & \textbf{\textcolor{blue} {+0.6}} & \textbf{\textcolor{blue} {+3.4}} & \textbf{\textcolor{blue} {+2.6}} & \textbf{\textcolor{blue} {+1.8}} & \textbf{\textcolor{blue} {+2.6}} & \textbf{\textcolor{blue} {+2.6}} & \textcolor{gray} {\ -0.2} & \textbf{\textcolor{gray} {+1.0}} \\
 \hline
 \textbf{MSTCN} & -                                 & 65.1 & 37.7 & 47.7 & 53.3 & 38.7 & 44.8 & \textcolor{gray}{67.7} & \textcolor{gray}{57.9}\\
 + CB~\cite{cui2019class}  & reweight               & \textcolor{red} {\ -1.0}  & \textcolor{blue} {+1.6}   & \textcolor{blue} {+1.1}   & \textcolor{blue} {+0.8}   & \textcolor{blue} {+0.7} & \textcolor{blue} {+0.8} & \textcolor{gray} {\ -0.3} & \textcolor{gray} {\ 0.0}\\
 + Focal~\cite{lin2017focal} & reweight             & \textcolor{blue} {+1.0}   & \textcolor{red} {\ -1.6}  & \textcolor{red} {\ -1.0}  & \textcolor{blue} {+0.3}   & \textcolor{red} {\ -0.7}& \textcolor{red} {\ -0.4} & \textcolor{gray} {+0.8} & \textcolor{gray} {\ -0.4}\\
 + BAGS~\cite{li2020overcoming} & ensemble          & \textcolor{blue} {+0.9}   & \textcolor{blue} {+1.8}   & \textcolor{blue} {+1.7}   & \textcolor{blue} {+4.4}   & \textcolor{blue} {+1.5} & \textcolor{blue} {+2.6} & \textcolor{gray} {+0.8} & \textcolor{gray} {+1.9}\\
 + $\tau$-norm~\cite{kang2019decoupling} & post-hoc & \textcolor{blue} {+0.2}   & \textcolor{red} {\ -1.5}  & \textcolor{red} {\ -1.1}  & \textcolor{red} {\ -0.6}  & \textcolor{red} {\ -1.3} & \textcolor{red} {\ -1.0} & \textcolor{gray} {+0.2} & \textcolor{gray} {\ -0.9}\\
 + LA~\cite{menon2020long} & logit adj.             & \textcolor{red} {\ -0.7}  & \textcolor{blue} {+2.4}   & \textcolor{blue} {+2.1}   & \textcolor{blue} {+2.7}   & \ 0.0 & \textcolor{blue} {+0.9} & \textcolor{gray} {\ -0.1} & \textcolor{gray} {\ 0.0}\\
 + LDAM~\cite{cao2019learning} & logit adj.         & \textcolor{blue} {+0.7}   & \textcolor{blue} {+0.1}   & \textcolor{blue} {+0.3}   & \textcolor{blue} {+0.7}   & \textcolor{blue} {+0.8} & \textcolor{blue} {+0.8} & \textcolor{gray} {\ -0.2} & \textcolor{gray} {+0.2}\\
 + Seesaw~\cite{wang2021seesaw} & logit adj.        & \textcolor{blue} {+1.2}   & \textcolor{blue} {+2.5}   & \textcolor{blue} {+2.4}   & \textcolor{blue} {+0.7}   & \textcolor{blue} {+0.3} & \textcolor{blue} {+0.5} & \textcolor{gray} {+0.9} & \textcolor{gray} {\ -0.1}\\
 + G-TLA(ours) & logit adj. & \textbf{\textcolor{blue} {+2.5}} & \textbf{\textcolor{blue} {+5.3}} & \textbf{\textcolor{blue} {+5.0}} & \textbf{\textcolor{blue} {+6.8}} & \textbf{\textcolor{blue} {+6.5}} & \textbf{\textcolor{blue} {+6.7}} & \textbf{\textcolor{gray} {+2.6}} & \textbf{\textcolor{gray} {+5.0}}\\
\hline
\end{tabular}}}
\label{tab:summary1}
\end{table}

\noindent \textbf{Evaluation metrics.} 
Temporal segmentation is traditionally evaluated \cite{farha2019ms, singhania2021coarse,yi2021asformer,wang2020boundary} using Mean over Frames accuracy for frames, Edit distance and F1-score at various IoU thresholds~$(0.10,0.25,0.50)$ for segments. However, these metrics aggregate globally across all samples, masking tail class performance. To assess each class effectively, we use balanced per-class recall for frame-wise and per-class F1$@0.25$ score for segment-wise evaluations, following long-tailed works~\cite{wang2021seesaw,tan2020equalization,kang2019decoupling}. The balanced metrics ensure that performance is not largely driven by the head classes. We compute the average performance within the head and tail groups and their harmonic mean to reflect the balance in our predictions accurately. See Suppl. for the F1 score at other thresholds and global metrics.

\subsection{Benchmark Comparisons}
We compare our work against seven long-tail methods across different datasets and backbones in \cref{tab:summary2} on YouTube \& \cref{tab:summary1} on Breakfast. We report per-class accuracy and F1 scores at IoU threshold 25\% for Head and Tail groups, respectively, and their harmonic mean (Hmean) to ensure that the models perform well across all classes, not just the head or tail. Some methods, like CB~\cite{cui2019class}, enhance tail group classes at the expense of head group classes (see \cref{tab:summary2}), while others, like Focal \cite{lin2017focal}, primarily enhance head classes with little improvement for tail classes (see \cref{tab:summary1}). Furthermore, some methods result in drops in segment-wise metric, $F1@25$, despite improved frame accuracy, indicating over-segmentation. For example, LA~\cite{menon2020long} and Seesaw~\cite{wang2021seesaw} in \cref{tab:summary2} sacrifice head class F1 scores despite achieving promising frame accuracy gains. We also observe that ensemble methods generally perform best across the long-tail method types, followed by LA, reweighting, and post-hoc methods.

Our method, G-TLA, demonstrates strong performance across different class groups, metrics, datasets, and backbones, effectively addressing the long-tail problem while minimizing over-segmentation. Using the Asformer backbone, our method outperforms the next best model by 2.2\% on YouTube and 1.5\% on Breakfast on frame accuracy, and by 2.5\% and 1.0\% on F1@25, respectively. When using the MSTCN backbone, we outperform the next best model by 4.1\% on YouTube and 2.6\% on Breakfast on frame accuracy, and by 1.7\% and 4.1\% on F1@25, respectively. Our method not only enhances frame accuracy but also consistently improves F1@25. Our method is further tested with the SOTA DiffAct~\cite{liu2023diffusion} backbone on Breakfast(\cref{tab:diff_reb}) to verify its effectiveness, improving per-class performance without compromising overall performance.
In summary, we find that directly applying standard long-tail models to temporal segmentation yields subpar performance, underscoring the significance of our method specifically designed to address the challenges posed by procedural activities.
\begin{table}[htb]
\caption{Additional results with DiffAct on Breakfast.}
\centering 
\setlength{\tabcolsep}{1.6mm}{
\resizebox{0.85\columnwidth}{!}{
\begin{tabular}{l|ccc|ccc|cc}
\hline
 \multirow{2}{*}{\textbf{Model}} & \multicolumn{3}{c|}{\textbf{Frame acc}} & \multicolumn{3}{c|}{\textbf{Segment F1@25}} & \multicolumn{2}{c}{\textcolor{gray}{\textbf{Global}}}  \\
\cline{2-9} & \textbf{Head} & \textbf{Tail} & \textbf{Hmean} & \textbf{Head} & \textbf{Tail} & \textbf{Hmean} & \textcolor{gray}{\textbf{Acc}} & \textcolor{gray}{\textbf{F1@25}}\\
 \hline
 \textbf{DiffAct}         & 74.8 & 42.6 & 54.3 & 77.5 & 49.2 & 60.2 & \textcolor{gray}{76.5} & \textcolor{gray}{75.3}\\
 + CB~\cite{cui2019class} & 74.1 & 43.3 & 54.7 & 77.5 & 49.1 & 60.2 & \textcolor{gray}{75.7} & \textcolor{gray}{75.8} \\
 + LA~\cite{menon2020long} & 75.3 & 44.5 & 55.9 & 78.1 & 49.0 & 60.2 & \textcolor{gray}{76.3} & \textcolor{gray}{75.6} \\
 + Seesaw~\cite{wang2021seesaw} & 73.2 & 38.9 & 50.8 & 76.0 & 44.0 & 55.8 & \textcolor{gray}{73.6} & \textcolor{gray}{72.0} \\
 + G-TLA(ours) & \textbf{75.6} & \textbf{45.5} & \textbf{56.8} & \textbf{78.4} & \textbf{50.4} & \textbf{61.4} & \textbf{\textcolor{gray}{76.6}} & \textbf{\textcolor{gray}{76.1}}\\
 \hline
\end{tabular}}}
\label{tab:diff_reb}
\end{table}

In \cref{tab:add_sum}, we present our model's performance on a recent long-tailed dataset Assembly101~\cite{sener2022}, along with two commonly used datasets, GTEA~\cite{fathi2011learning} and 50Salads~\cite{stein2013combining}, which have smaller vocabulary sizes and are less imbalanced. For GTEA and Assembly101, we apply clustering to determine the group for each sequence (3 groups for GTEA and 2 groups for Assembly101), with each group containing several activities. For 50Salads, which contains a single activity, we forgo the group-wise framework and exclusively implement temporal logit adjustment. Our method demonstrates competitive results over the baseline by a large margin, further validating its effectiveness. Notably, the tail performance is even higher than the head for GTEA, emphasizing the lack of imbalance.
 
\begin{table}[htb]
\caption{Additional results on 50salads, GTEA, and Assembly101.}
\centering 
\setlength{\tabcolsep}{1.6mm}{
\resizebox{0.95\columnwidth}{!}{
\begin{tabular}{l|l|lll|lll|ll}
\hline
\multirow{2}{*}{\textbf{Dataset}} & \multirow{2}{*}{\textbf{Model}} & \multicolumn{3}{c|}{\textbf{Frame acc}} & \multicolumn{3}{c|}{\textbf{Segment F1@25}} & \multicolumn{2}{c}{\textbf{\textcolor{gray}{Global}}} \\
\cline{3-10} & & \textbf{Head} & \textbf{Tail} & \textbf{Hmean} & \textbf{Head} & \textbf{Tail} & \textbf{Hmean}  & \textbf{\textcolor{gray}{Acc}} & \textbf{\textcolor{gray}{F1@25}} \\
 \hline
 \multirow{4}{*}{\textbf{50salads}} &\textbf{AsFormer} & 90.6 & 77.4 & 83.5 & 87.5 & 80.3 & 83.8 & \textcolor{gray}{85.2} & \textcolor{gray}{82.3} \\
 & + G-TLA(ours) & \textbf{90.8} & \textbf{79.7} & \textbf{84.9} & \textbf{89.4} & \textbf{83.1} & \textbf{86.1} & \textbf{\textcolor{gray}{86.3}} & \textbf{\textcolor{gray}{84.6}}\\
\cline{2-10} &\textbf{MSTCN} & 87.7 & 70.0 & 77.9 & 85.7 & 72.1 & 78.3 & \textcolor{gray}{81.1} & \textcolor{gray}{75.9}\\
 & + G-TLA(ours) & \textbf{89.0} & \textbf{71.7} & \textbf{79.4} & \textbf{86.8} & \textbf{73.8} & \textbf{79.8} & \textbf{\textcolor{gray}{81.9}} & \textbf{\textcolor{gray}{77.3}}\\
 \hline
 \hline
 \multirow{4}{*}{\textbf{GTEA}} &\textbf{AsFormer} & \textbf{80.6} & 81.7 & 81.2 & \textbf{72.5} & 85.4 & 78.4 & \textcolor{gray}{81.1} & \textcolor{gray}{89.4}\\
 & + G-TLA(ours) & 80.2 & \textbf{84.5} & \textbf{82.3} & 72.0 & \textbf{90.4} & \textbf{80.2} & \textbf{\textcolor{gray}{81.2}} & \textbf{\textcolor{gray}{89.6}}\\
\cline{2-10} &\textbf{MSTCN} & \textbf{77.6} & 80.3 & 78.9 & 69.4 & 86.6 & 77.0 & \textcolor{gray}{78.0} & \textcolor{gray}{87.2}\\
 & + G-TLA(ours) & 77.5 & \textbf{83.7} & \textbf{80.5} & \textbf{69.5} & \textbf{90.3} & \textbf{78.5} & \textbf{\textcolor{gray}{78.6}} & \textbf{\textcolor{gray}{87.9}}\\
\hline
 \hline
 \multirow{4}{*}{\textbf{Assembly101}} &\textbf{AsFormer} & 35.2 & 5.7 & 9.8 & 29.0 & 4.8 & 8.2 & \textbf{\textcolor{gray}{41.1}} & \textbf{\textcolor{gray}{30.4}}\\
 & + G-TLA(ours) & \textbf{36.8} &\textbf{9.2} & \textbf{14.7} & \textbf{30.7}& \textbf{8.3} & \textbf{13.1} & \textcolor{gray}{41.0} & \textcolor{gray}{29.8}\\
 \cline{2-10} &\textbf{MSTCN} & 33.9 & 4.7 & 8.2 & 26.3 & 3.9 & 6.8 & \textbf{\textcolor{gray}{39.8}} & \textcolor{gray}{27.2}\\
 & + G-TLA(ours) & \textbf{34.9} & \textbf{8.0}& \textbf{13.0} & \textbf{30.2} & \textbf{5.8} & \textbf{9.7} & \textcolor{gray}{39.2} & \textbf{\textcolor{gray}{28.5}}\\
\hline
\end{tabular}}}
\label{tab:add_sum}
\end{table}

\begin{table}[htb]
\caption{Ablate group classification(GP), logit adjustment(LA), temporal factor(TF).} 
\centering 
\setlength{\tabcolsep}{1.5mm}{
\resizebox{0.98\columnwidth}{!}{
\begin{tabular}{ccc|ccc|ccc||ccc|ccc}
\hline
\multirow{3}{*}{\textbf{GP}} & \multirow{3}{*}{\textbf{LA}} & \multirow{3}{*}{\textbf{TF}} & \multicolumn{6}{c||}{\textbf{MSTCN}} & \multicolumn{6}{c}{\textbf{ASFormer}} \\ 
\cline{4-15} & & & \multicolumn{3}{c|}{\textbf{Frame acc}} & \multicolumn{3}{c||}{\textbf{Seg. F1}} & \multicolumn{3}{c|}{\textbf{Frame acc}} & \multicolumn{3}{c}{\textbf{Seg. F1}}\\
\cline{4-15} & & & Head & Tail & Hmean & Head & Tail & Hmean & Head & Tail & Hmean & Head & Tail & Hmean\\
\hline
\xmark & \xmark & \xmark & 65.1 & 37.7 & 47.7 & 53.3 & 38.7 & 44.8 & 69.7 & 39.8 & 50.7 & 69.9 & 43.9 & 53.9 \\ 
\xmark & \cmark & \xmark & 64.4 & 40.1 & 49.8 & 56.0 & 38.7 & 45.7 & 70.1 & 40.4 & 51.3 & 71.2 & 44.7 & 54.9 \\ 
\xmark & \cmark & \cmark & 65.7 & 41.3 & 50.7 & 56.3 & 38.9 & 46.0 & 69.4 & 41.7 & 52.2 & 70.7 & 44.8 & 54.9 \\ 
 \cmark & \xmark & \xmark & 67.5 & 40.8 & 50.9 & \textbf{60.3} & 44.8 & 51.3 & 69.7 & 40.8 & 51.5 & 71.1 & 44.9 & 55.0 \\ 
 \cmark & \cmark & \xmark & 66.5 & 41.8 & 51.3 & 59.9 & 44.7 & 51.2 & 70.2 & 41.4 & 52.1 & 71.4 & 44.7 & 55.0 \\ 
 \cmark & \cmark & \cmark & \textbf{67.6} & \textbf{43.0} & \textbf{52.7} & 60.1 & \textbf{45.2} & \textbf{51.5} & \textbf{70.3} & \textbf{43.2} & \textbf{53.3} & \textbf{71.7} & \textbf{46.5} & \textbf{56.5} \\ 
\hline
\end{tabular}}
}
\label{tab:ab_comb}
\end{table}

\subsection{Ablation Studies} 
We present ablations on Breakfast; please see the Suppl. for other datasets. 

\noindent \textbf{G-TLA Components.} 
We assess the contributions of group-wise classification (GP), naive logit adjustment (LA), and temporal factors (TF) in \cref{tab:ab_comb}. GP significantly reduces over-segmentation by 5.5\% with MSTCN and 1.1\% for ASFormer. Naive LA improves tail accuracy by 2.4\% but decreases head accuracy by 0.7\% for MSTCN. Temporal priors help reduce false positives that violate the ordering prior, further improving LA. Combining all our components within G-TLA achieves a balanced result in both frame and segment metrics. 

\begin{table}[t]
 \begin{minipage}{.48\linewidth}
 \caption{Varying $\eta$ for group-wise classification, with fixed number of groups $n=10$ and $\tau=0.5$ on MSTCN}
 \label{tab:eta}
\centering 
\setlength{\tabcolsep}{1.1mm}{
\resizebox{0.9\columnwidth}{!}{
 \begin{tabular}{c|ccc|ccc}
 \hline
 \multirow{2}{*}{$\eta$} & \multicolumn{3}{c|}{\textbf{Frame acc}} & \multicolumn{3}{c}{\textbf{Seg. F1}} \\ 
 \cline{2-7} & Head & Tail & Hmean & Head & Tail & Hmean \\ \hline
 0.1 & 67.3 & 41.0 & 51.0 & 60.4 & 44.0 & 50.9 \\ 
 0.3 & 67.2 & 42.1 & 51.8 & 60.5 & 44.8 & 51.5 \\ 
 0.5 & \textbf{67.6} & \textbf{43.0} & \textbf{52.7} & 60.1 & 45.2 & 51.5 \\
 0.7 & 67.5 & 42.3 & 52.1 & \textbf{61.1} & \textbf{45.7} & \textbf{52.2} \\ 
 \hline
 \end{tabular}}}
 \end{minipage}%
 \quad
 \begin{minipage}{.48\linewidth}
 \centering
 \caption{Varying $\tau$ for temporal logit adjustment, with fixed number of groups $n=10$ and $\eta=0.5$ on MSTCN}
 \label{tab:rau}
\centering 
\setlength{\tabcolsep}{1.1mm}{
\resizebox{0.9\columnwidth}{!}{
 \begin{tabular}{c|ccc|ccc}
 \hline
 \multirow{2}{*}{$\tau$} & \multicolumn{3}{c|}{\textbf{Frame acc}} & \multicolumn{3}{c}{\textbf{Seg. F1}} \\ 
 \cline{2-7} & Head & Tail & Hmean & Head & Tail & Hmean \\ \hline
 0.1 & 67.4 & 40.8 & 50.9 & 60.9 & 43.3 & 50.6 \\ 
 0.3 & 67.5 & 41.8 & 51.6 & \textbf{61.0} & 43.2 & 50.6 \\ 
 0.5 & \textbf{67.6} & \textbf{43.0} & \textbf{52.7} & 60.1 & \textbf{45.2} & \textbf{51.5} \\
 0.7 & \textbf{67.6} & 42.4 & 52.1 & 60.6 & 43.9 & 50.9 \\ 
 \hline
 \end{tabular}}}
 \end{minipage} 
\end{table}

\begin{figure}[tb]
 \centering
 \subfloat[][MSTCN.]{\includegraphics[scale=0.17]{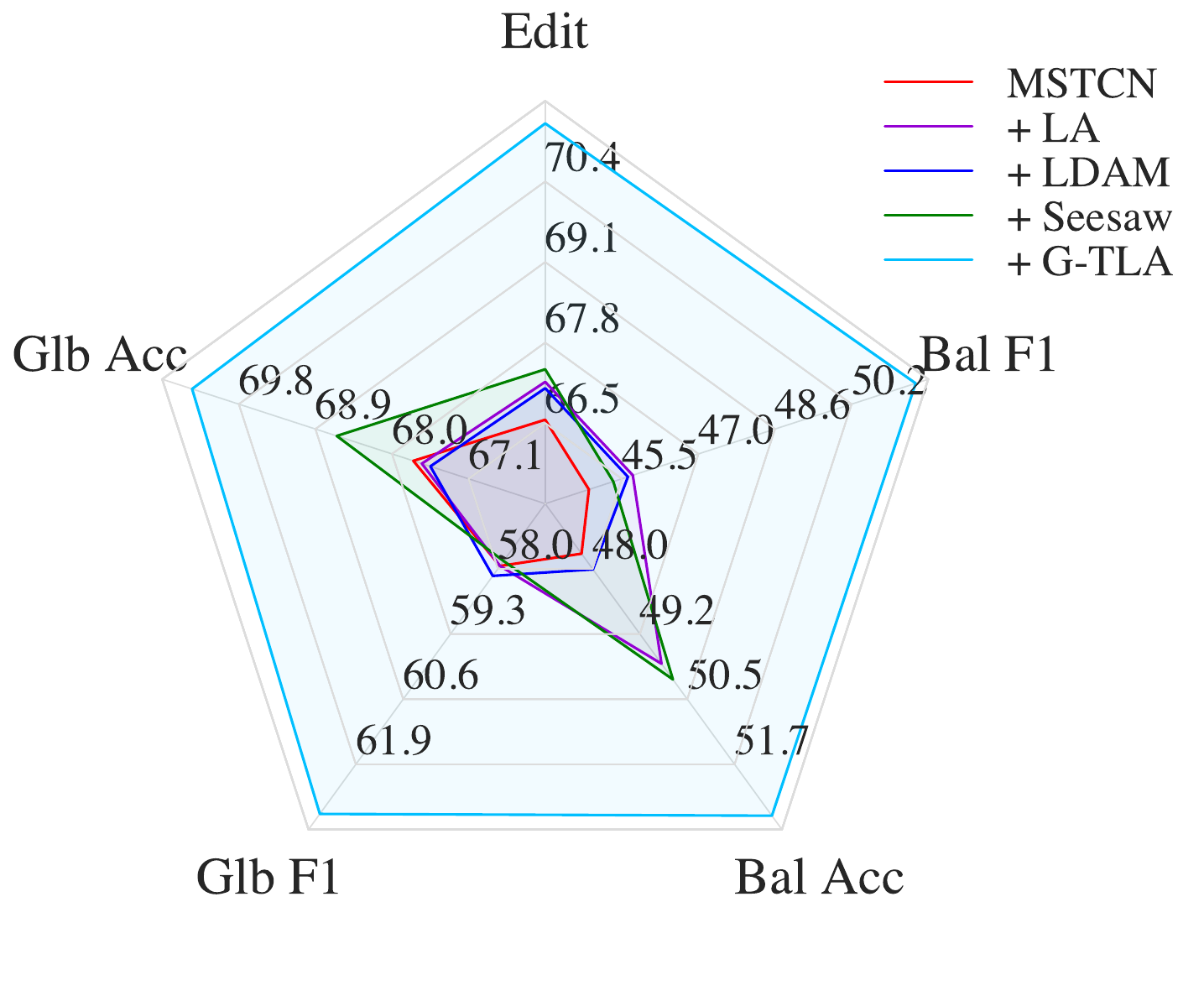}}
 \quad
 \subfloat[][AsFormer.]{\includegraphics[scale=0.17]{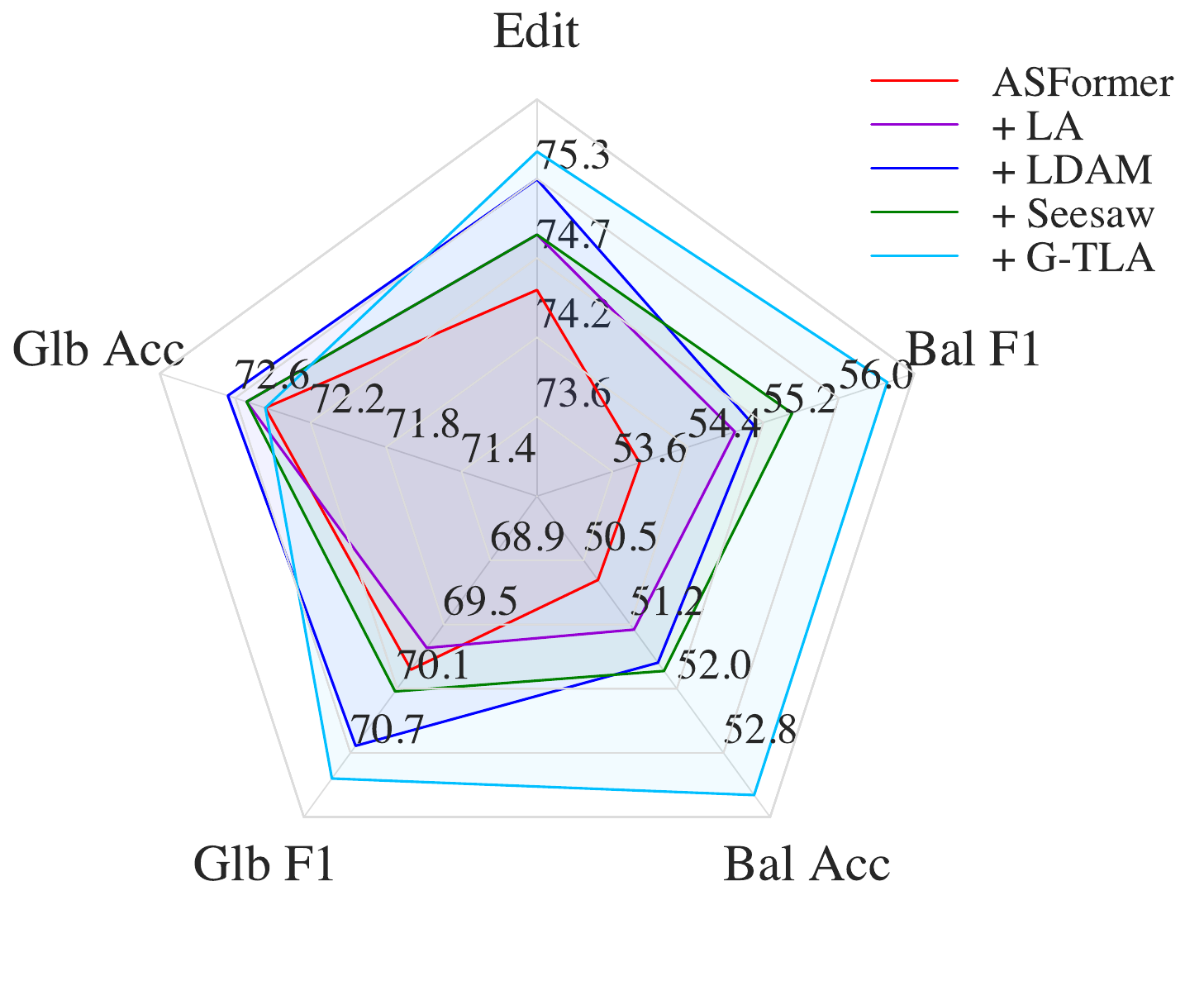}}
 \caption{Radar charts of different logit adjustment methods, measuring the performance along balanced and global metrics on Breakfast with MSTCN and AsFormer.}
 \label{fig:radar_mstcn}
\end{figure}

\begin{figure}[t]
 \centering
 {\includegraphics[scale=0.58]{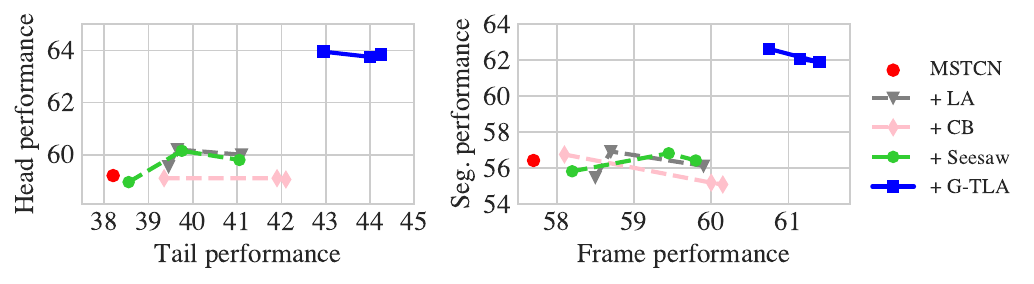}}
 \caption{Head-Tail \& Frame-Segment trade-offs on Breakfast with MSTCN. }
 \label{fig:tradeoff}
\end{figure}

\noindent \textbf{$\eta$ for group-wise classification.}
\cref{tab:eta} shows the impact of hyperparameter $\eta$ in ~\cref{eq: glta_loss}. A small $\eta$ reduces suppression of tail classes by down-weighting negative gradients from the \emph{`others'} class, but if too small, it harms group identification during inference. Conversely, a large $\eta$ over-emphasizes the \emph{`others'} class, harming tail performance. The results suggest an optimal value of $\eta$ is 0.5.

\noindent \textbf{$\tau$ for temporal logit adjustment.} A small $\tau$ in \cref{eq: la_tas} represents minimal adjustment, resulting in less improvement for tail classes in ~\cref{tab:rau}. Conversely, a large $\tau$ biases towards tail classes and introduces more false positives, negatively impacting head and segment-wise performance. Our experiments show that $\tau = 0.5$ achieves optimal performance.

\subsection{ Analyzing the effects of G-TLA}
\noindent \textbf{Individual metrics.} \cref{fig:radar_mstcn} visualizes various methods' global and per-class results with radar charts, including global accuracy, F1 score, and Edit score, as well as the harmonic mean of balanced accuracy and F1@25 score. Our method is significantly more balanced, as indicated by the largest enclosed area. In particular, our method excels in segment-wise performance, including Edit score and global \& balanced F1 score, demonstrating our approach's effectiveness in reducing over-segmentation while enhancing balanced accuracy.

\begin{figure*}[htb]
\centering
\begin{minipage}[b]{.61\textwidth}
 \centering
 \subfloat[][\emph{"making cereals"} activity on Breakfast using MSTCN.]{\includegraphics[width=7.3cm,height=1.7cm]{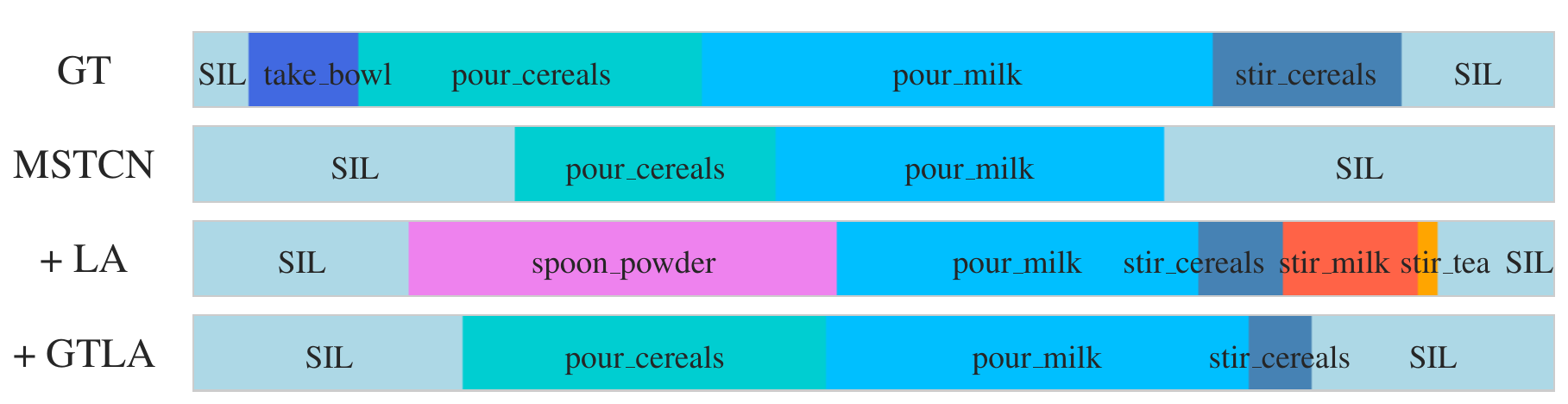}}
 \hfill
 \subfloat[][\emph{"changing tire"} activity on Youtube using MSTCN.]{\includegraphics[width=7.5cm,height=1.7cm]{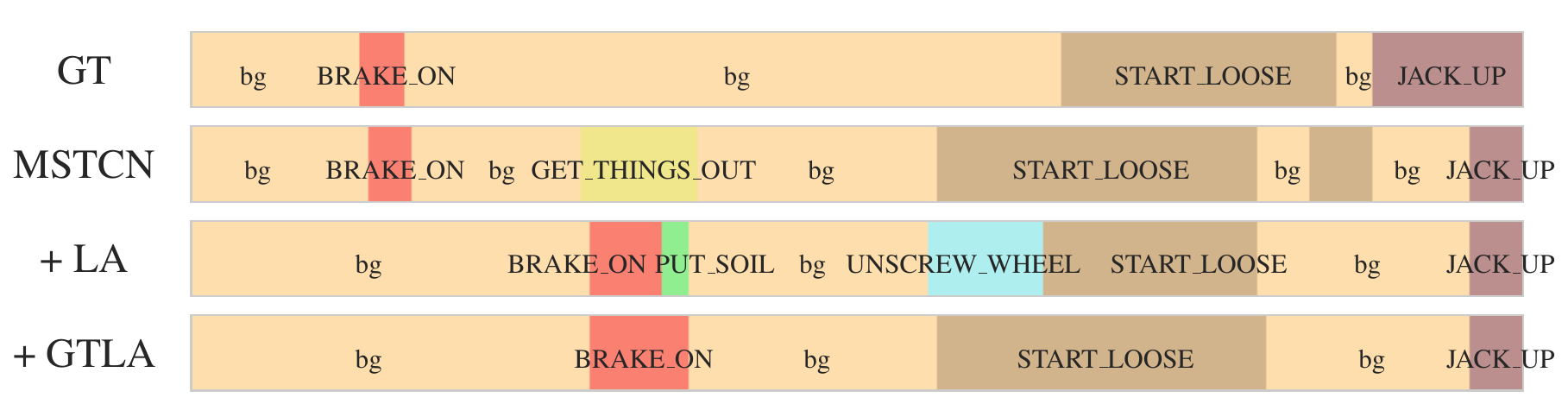}}
 \caption{G-TLA effectively reduces activity-irrelevant predictions, \eg actions \emph{`spoon powder'} and \emph{`stir milk'} on Breakfast, and \emph{`PUT SOIL'} on Youtube. G-TLA also mitigates predictions that violate temporal priors, \eg \emph{`UNSCREEW WHEEL'} occurring before \emph{`START LOOSE'} on Youtube. }
 \label{fig:quality}
\end{minipage}
\ \
\begin{minipage}[b]{.35\textwidth}
 \centering
 \includegraphics[scale=0.62]{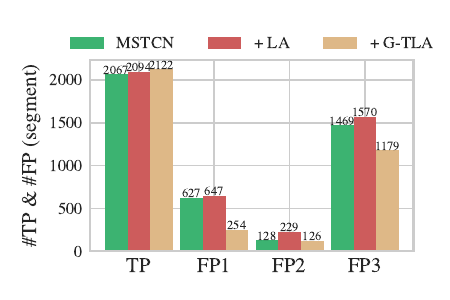}
 \caption{Distribution of True (TP) and False Positives (FP), categorized into: FP1 (activity-irrelevant), FP2 (activity-relevant with ordering violations), and FP3 (activity-relevant and following action ordering derived from temporal bounds).
 }
 \label{fig:fps}
\end{minipage}%
\end{figure*}

\noindent \textbf{Trade-off trends. } Long-tail temporal action segmentation exhibits two distinct trade-offs. First is a head-tail trade-off, where the head is impacted negatively when the tail is improved. Second is a frame-segment trade-off, where enhancing the tail improves frame metrics but hurts segment metrics. The extent of these trade-offs is directly influenced by the hyperparameters of respective long-tail methods, as shown in \cref{fig:tradeoff}. Head performance involves head group metrics of per-class accuracy and F1 score, and tail performance involves tail group metrics. Frame performance measures the average of global and per-class accuracy, while segment performance involves Edit score, global F1@25, and per-class F1@25. As the plot indicates, our method achieves a much better balance than others. The curve of our method consistently remains above others, emphasizing the boost of tail classes. This contributes to improved balanced metrics while maintaining competitive head and segment performance. See the Supplementary for hyperparameters and trends on other datasets and backbones.

\noindent \textbf{Qualitative \& quantitative results. } 
We qualitatively compare the predictions of logit adjustment methods. \cref{fig:quality} shows the output of MSTCN on Breakfast and YouTube, revealing the common issue of introducing activity-irrelevant classes when enhancing tail classes. For instance, emphasizing the tail class \emph{'stir milk'} on Breakfast introduces false positives in unrelated activity \emph{"making cereal"}. 
Additionally, existing methods ignore action ordering, resulting in false positives that violate temporal priors, \eg on Youtube, action \emph{`UNSCREW WHEEL'} is wrongly predicted before \emph{`START LOOSE'} for activity \emph{"changing tire"}. Our proposed method effectively addresses both types of false positives, enhancing the overall prediction logic. We also report the quantitative results on Breakfast with MSTCN in \cref{fig:fps}. Our method reduces the number of segments for activity-irrelevant(FP1) and ordering-violated(FP2) false positives, mitigating over-segmentation. Meanwhile, it also helps to increase true positives(TP) and reduce false positives that are both activity-relevant and ordering-valid(FP3), mainly due to the group-wise classification.

\noindent \textbf{Group identification accuracy.}  Group identification is crucial and impacts the final performance. Our group-wise classification improves activity identification. For instance, we achieve 90.1\% accuracy compared to 87.2\% of the MSTCN baseline on Breakfast. More details and results are in the Supplementary.

%% file: sec/6_conclusion.tex
\section{Conclusion}
\label{sec:conclusion}

This paper targets long-tail temporal action segmentation, addressing challenges from temporal class correlations and performance trade-offs of head \& tail classes at frame- \& segment-level. Our proposed Group-wise Temporal Logit Adjustment (G-TLA) scheme integrates video activity labels and action order priors to capture class inter-dependencies, enhancing balanced performance while maintaining global performance. Future efforts would prioritize highly imbalanced datasets, considering both long-tail and few-shot learning scenarios.

%% file: sec/7_suppl.tex
\section*{Supplementary Material}
\label{sec:suppl}

\subsection*{A. \quad Long-tail Temporal Action Segmentation}
\begin{figure}[htb]
  \centering
  \includegraphics[height=4.7cm]{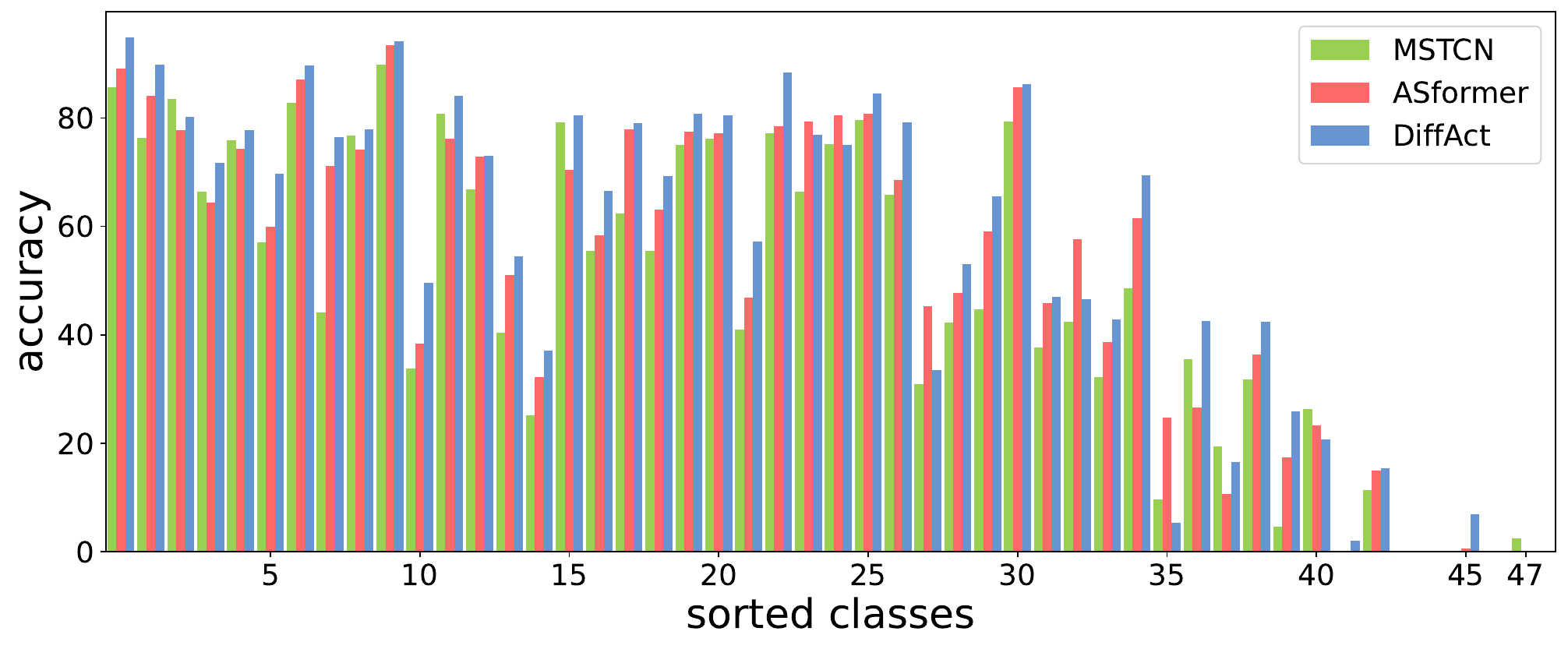}
  \caption{Class-wise accuracy distribution on Breakfast with MSTCN, ASFormer and DiffAct. Classes are sorted by their frame counts in the training set. }
  \label{fig:acc_bf_supp}
\end{figure}

The long tail problem has been overlooked in temporal action segmentation. 
State-of-the-art methods~\cite{farha2019ms, yi2021asformer, liu2023diffusion} perform poorly and often do not predict any tail classes correctly. For example, MSTCN~\cite{farha2019ms}, ASFormer~\cite{yi2021asformer}, and DiffAct~\cite{liu2023diffusion} how zero accuracies for 5, 5, and 4 out of 48 classes on the Breakfast dataset as shown in the per-class accuracy plot in \cref{fig:acc_bf_supp}. Our paper is the first to address the long tail problem in temporal action segmentation.

\subsection*{B. \quad Grouping Without Activity Labels} 
During training, our group-wise classification relies on grouping classes based on activity labels. However, in scenarios where the activity label is unavailable or should not be utilized, an alternative approach is forming groups through sequence clustering. Notably, clustering results often align with the underlying activity label. Some group examples are shown in \cref{tab:gp_eg_br}. During testing, we use \cref{eq: group_infer} to identifyied groups. Details of the group identification results can be found in Sec. E.

\begin{table}[h]
\caption{Grouping examples. }
\centering
\small
\begin{tabular}{c|c|c}
\hline
\textbf{Dataset} & \textbf{Group} & \textbf{Action Classes} \\ 
\hline
 Breakfast & \emph{``coffee''} & \makecell{ 
\emph{take\_cup}, \emph{pour\_sugar}, \emph{spoon\_sugar}, \emph{SIL}, \\ 
 \emph{pour\_coffee}, \emph{stir\_coffee}, \emph{pour\_milk}} \\ 
\hline 
GTEA & \emph{``Cluster 1''} & \makecell{\emph{Pour}, \emph{Take}, \emph{Close}, \emph{Put}, \emph{Open}, \emph{Fold}, \emph{Background}} \\ 
\hline 
\end{tabular}
\label{tab:gp_eg_br}
\end{table}

\subsubsection{Clustering algorithm}
We cluster sequences according to the action frequency distribution, leveraging its ability to capture the action co-occurrence patterns~\cite{ding2022leveraging}. Given a sequence $i$, the action frequency distribution $q$ is defined as the normalized occurrence of frames for all the actions:
\begin{equation}
     q_i(c) = \frac{1}{T_i} \sum_t^{T_i} \mathbb{1}(y_t = c), \ \ c \in[1, \cdots L].
\label{eq: seq_rep}
\end{equation}

Then, we can define the distance criterion between two sequences $i, j$ using the Kullback-Leibler(KL)-divergence:
\begin{equation}
     dist(i,j) = \frac{1}{2} \sum_c q_i(c) \log \frac{q_i(c)}{q_j(c)}  +  q_j(c) \log \frac{q_j(c)}{q_i(c)} 
\label{eq: seq_rep_kl}
\end{equation}
We average over the forward and backward KL-divergence to ensure symmetry in the distance measure. Based on the defined distance criterion, we apply hierarchical clustering~\cite{jolly2018machine} with a predefined number of groups and a tuned linkage criterion to establish the sequence-to-group mapping. 

\subsubsection{Effect of the number of groups $n$. } We present results of using clustering on Breakfast with MSTCN to assess the impact of the number of groups. Setting the number of clusters to $n=10$ yields the same groups as using the activity label. Further reducing $n$ leads to the merging of several activities. For instance, setting $n=8$ merges the activities \emph{"ceareal"}-\emph{"milk"} and \emph{"friedegg"}-\emph{"scrambledegg"}. According to the results in \cref{tab:ab_n_supp}, finer clustering, \ie more detailed separation of the activities, contributes to better performance by reducing the false positives from activity-irrelevent classes to a larger extent.

\begin{table}[ht]
\caption{Varying the number of groups $n$ for group-wise classification, with fixed $\eta =0.5$ and $\tau=0.5$}
\centering
\begin{tabular}{c|ccc|ccc}
 \hline
 \multirow{2}{*}{$n$} & \multicolumn{3}{c|}{\textbf{Frame acc}} & \multicolumn{3}{c}{\textbf{Seg. F1}} \\ 
 \cline{2-7} & Head & Tail & Hmean & Head & Tail & Hmean \\ \hline
 3 & 65.9 & 39.7 & 49.6 & 56.3 & 40.3 & 47.0 \\ 
 5 & 66.3 & 41.1 & 50.7 & \textbf{61.2} & 43.4 & 50.8 \\ 
 8 & 66.5 & 41.5 & 50.9 & 60.3 & 44.5 & 51.2 \\ 
 10 & \textbf{67.6} & \textbf{43.0} & \textbf{52.7} & 60.1 & \textbf{45.2} & \textbf{51.5} \\ 
 \hline
 \end{tabular}
\label{tab:ab_n_supp}
\end{table}

\subsection*{C. \quad Temporal Priors} 
For an action $c$, we defined two sets $S_{bf}[c]$ and $S_{af}[c]$ that contain actions that must precede and follow action $c$. These two sets are utilized to fix the temporal bounds when employing logit adjustment on action $c$. These two sets are exclusive and can be extracted from the training data. The algorithm for finding these two sets for a given class $c$ is given in \cref{algo:tem_supp}. Examples of extracted temporal bounds for actions on Breakfast can be found in \cref{tab:tem_ex_supp}.

\begin{algorithm}
\caption{Exacting Temporal Bounds for Class $c$}
\label{algo:tem_supp}
    \hspace*{\algorithmicindent} \textbf{Input:} \ Training sequence labels $\mathcal{Y}$, Class $c$ \\
    \hspace*{\algorithmicindent} \textbf{Initialize:} \ $A = \text{set()}$, $B = \text{set()}$
	\begin{algorithmic}[1]
        \For{$ Y \ \text{in} \ \mathcal{Y} $} \textcolor{magenta}  { \quad \emph{// for each sequence}}
            \If{$c \notin Y$}
                \State \text{continue}
            \EndIf
            \State $ls$ = \text{get\_seg\_label}$(Y)$ \quad  \textcolor{magenta}{\emph{// segment-wise label}}
            \State $ids$ = $ls$.index($c$)
            \For{$ i \in ids $} 
		      \State $ A = A \cup ls[i\text{+1}:]$ \quad \textcolor{magenta}{\emph{// update actions after $c$}}
                \State $ B = B \cup ls[:i]$ \quad \textcolor{magenta}{\emph{// update actions before $c$}}
		\EndFor  
        \EndFor
        \State $S_{bf}[c] = B - A$  \quad \textcolor{magenta}{\emph{// must precede $c$}}
        \State $S_{af}[c] = A - B$  \quad \textcolor{magenta}{\emph{// must follow $c$}}
	\end{algorithmic}  
    \hspace*{\algorithmicindent} \textbf{Return:} $S_{bf}[c], \ S_{af}[c] $
\end{algorithm} 

\begin{table}[ht]
\caption{Examples of extracted temporal bounds for actions in Breakfast dataset.}
\centering
\small
\begin{tabular}{c|c|c}
\hline
\textbf{Actions}  &  $S_{bf}[c]$  & $S_{af}[c]$   \\ \hline
\emph{pour cereals} & \emph{take bowl} & \emph{pour milk}, \emph{stir cereals} \\
\emph{take bowl}   & -     & \emph{pour milk}, \emph{pour cereals}, \emph{stir cereals} \\
\emph{add teabag}   & \emph{take cup}  & \emph{pour sugar}, \emph{spoon sugar}, \emph{stir tea}       \\
\emph{stir milk}    & \emph{take cup}  & -   \\                            
\hline                                                           
\end{tabular}
\label{tab:tem_ex_supp}
\end{table}

\subsection*{D. \quad Experimental Setting}
\subsubsection{Dataset.}
\begin{figure*}
\centering
    \begin{subfigure}{0.49\linewidth}
    \includegraphics[scale=0.14]{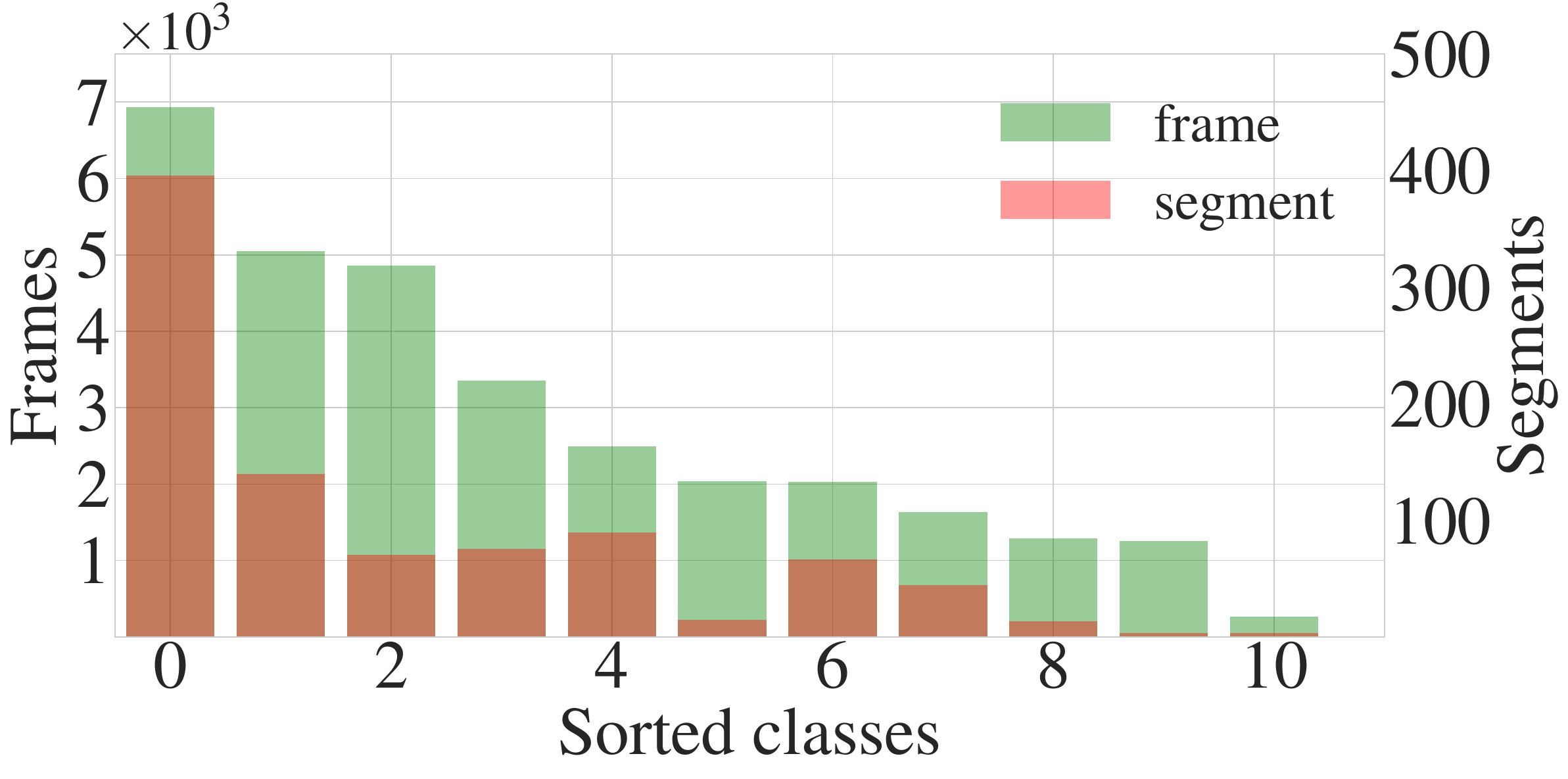}
    \caption{GTEA}
    \end{subfigure}
    \begin{subfigure}{0.49\linewidth}
    \includegraphics[scale=0.14]{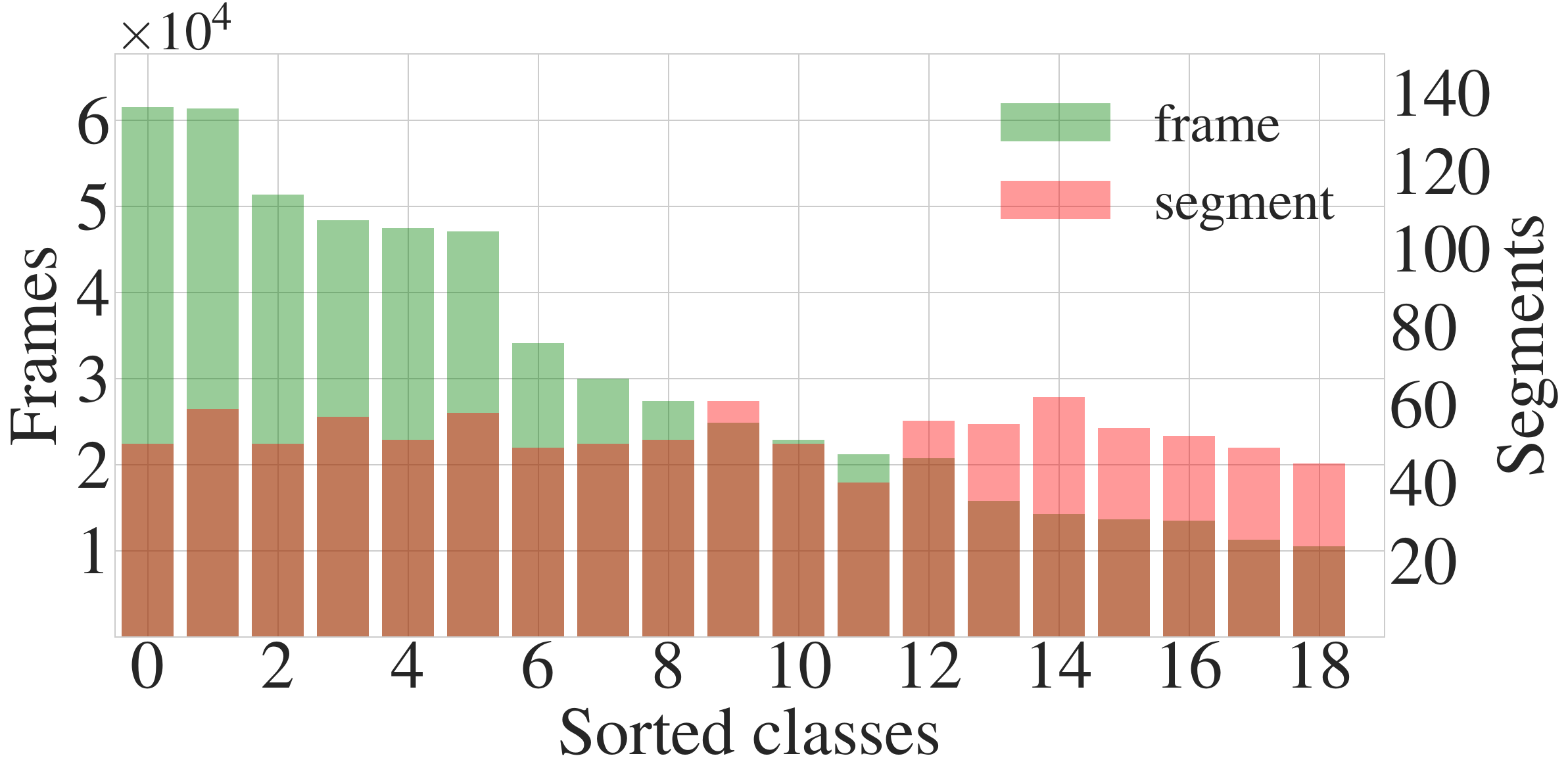}
    \caption{50salads}
    \end{subfigure}
    \begin{subfigure}{0.5\linewidth}
    \includegraphics[scale=0.15]{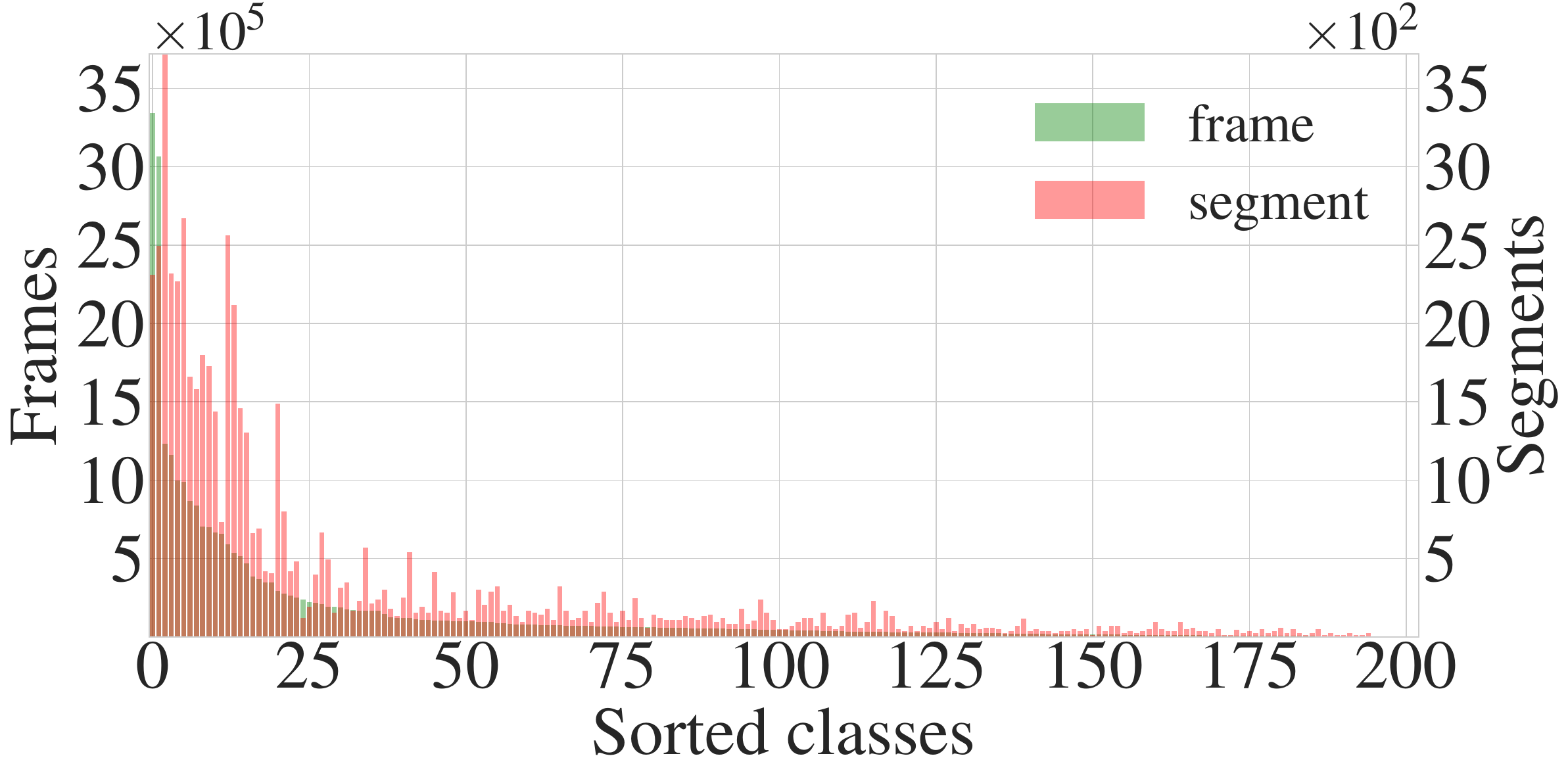}
    \caption{Assembly101}
    \end{subfigure}
    \caption{ Data distribution of Assembly101, GTEA, and 50salads. }
 \label{fig:data_dist_supp}
\end{figure*}

We conduct experiments on five benchmarks. (1) Breakfast consists of 1712 videos with ten video-level activities for making breakfast. On average, the videos are 2.3 minutes long with 48 action classes. (2) YouTube Instructional is a collected dataset that includes five instructional activities. It comprises 30 videos for each activity, totaling 46 unique action classes. (3) Assembly101 is a recently collected dataset for dissembling and assembling take-apart toys. It has a collection of 4321 videos with an average length of 7.1 minutes and 202 coarse actions. (4) GTEA contains 28 videos of seven procedural activities recorded in a single kitchen, with a total of 11 actions.  (5) 50Salads is composed of 50 recorded videos of making mixed salads involving 19 actions. Despite the less imbalanced nature, we include the GTEA and 50Salads results for comprehensive evaluation, as these datasets are widely utilized within the research community. Data distribution of these datasets is illustrated in \cref{fig:data_dist_supp}. 

\subsubsection{Implementation Details. }
The experimental configurations of base models are summarized in \cref{tab:base_set_supp}. All models are trained to reduce over-segmentation with an extra smoothing loss~\cite{farha2019ms} with $\lambda = 0.15$. We follow the protocols in original papers for any details not specified here. As MSTCN and ASFormer are multi-stage models, the long-tail methods are exclusively applied at the final stage. Applying these methods to all stages results in degraded performance.

\begin{table*}[ht]
  \caption{ Base model setup}
  \label{tab:base_set_supp}
  \centering
  \begin{tabular}{ccccccc}
    \toprule
      Model &  Optimizer  &  lr & weight decay &  Batch size & Epochs &  Sample rate \\
    \midrule
     MSTCN & Adam &  $5 \times 10^{-4}$ & - & 1 & 50 & 1   \\
     ASFormer & Adam &  $1 \times 10^{-4}$ & $1 \times 10^{-5}$ & 1 & 60 & 4   \\
    \bottomrule
  \end{tabular}
\end{table*}

\noindent We include several types of methods for comparison. 
\begin{itemize}
   \item \textbf{Re-weighting.} Focal~\cite{lin2017focal} assigns different weights to samples based on their difficulty; CB~\cite{cui2019class} calculates the weight for each class based on its effective number of samples. 
   \item  \textbf{Logit adjustment.} LA~\cite{menon2020long} and LDAM~\cite{cao2019learning} both adjust the logits based on the class prior, where the prior is estimated with the class-independent assumption.  Seesaw~\cite{wang2021seesaw} dynamically re-balances the gradients of positive and negative samples by adjusting the logits.
   \item \textbf{Post-hoc process.} $\tau$-norm~\cite{kang2019decoupling} normalizes the weights of a learned classifier to achieve a balanced classifier. 
   \item \textbf{Ensemble.} BAGS~\cite{li2020overcoming} utilizes group-wise training by modulating the training for head and tail classes separately to ensures both are sufficiently trained.
\end{itemize} 

Hyperparameters used in above methods are selected through grid search. Hyperparameters that yield the best overall balanced and global metrics results are selected. Our method adopts the hyperparameter $\eta$ to balance the target and non-target group losses, and $\tau$ as in LA~\cite{menon2020long} to tune the head-tail trade-off. \cref{tab:hyper_supp} gives the search space for different methods. Please refer to the references for the meaning of hyperparameters in each method.

\begin{table}[H]
  \caption{Search space of hyperparameters}
  \label{tab:hyper_supp}
  \centering
  \begin{tabular}{cc}
    \toprule
    Method & Search space \\   
    \midrule
     Focal~\cite{lin2017focal} & $\gamma \in \{0.5, 1.0, 1.5\}$ \\
     CB~\cite{cui2019class} & $\beta \in \{0.9, 0.99, 0.999\}$   \\
     LA~\cite{menon2020long}  & $\tau \in \{0.1, 0.3, 0.5, 0.7\}$   \\
     LDAM~\cite{cao2019learning}  & $s \in \{1, 3, 5, 10\}$, $m \in \{0.5, 1.0, 1.5\}$   \\
     Seesaw~\cite{wang2021seesaw}  & $p \in \{0.1, 0.2, 0.4\}$, $q \in \{0.5, 1.0, 1.5\}$    \\
     $\tau$-norm~\cite{kang2019decoupling}  & $\tau \in \{ 0.5, 1.0, 1.5\}$    \\
     BAGS~\cite{li2020overcoming} & $\beta \in \{ 2, 4, 8\}$, $N \in \{ 2, 3, 4\}$     \\
     G-TLA(ours)  & $\eta \in \{0.1, 0.3, 0.5, 1.0\}$    \\
    \bottomrule
  \end{tabular}
\end{table}

The hyperparameters used for each dataset, backbone, and method are detailed in Table \ref{tab:hyper_break_supp}. We omit $\tau$-norm in the tables as the results always favor $\tau=1.0$ for $\tau$-norm. Our approach adopts a group-wise classification framework, where the number of groups is decided either based on activity label or by clustering: for Breakfast and Youtube, we group based on activity labels with $n$ set to 10 and 5, respectively; for GTEA and Assembly, clustering forms groups with $n$ set to 3 and 2, respectively; for 50salads, group-wise classification is discarded, equivalent to $n=1$. 

\begin{table*}[htb]
  \caption{Hyperparameters for different experimental settings. }
  \label{tab:hyper_break_supp}
  \centering
  \resizebox{0.93\columnwidth}{!}{
  \begin{tabular}{ccccccccc}
    \toprule
    \multirow{2}{*}{Data} & \multirow{2}{*}{Model} & Focal~\cite{lin2017focal} & CB~\cite{cui2019class} & LA~\cite{menon2020long} & LDAM~\cite{cao2019learning} & Seesaw~\cite{wang2021seesaw} &  BAGS~\cite{li2020overcoming} &  G-TLA \\ 
    & & $\gamma$ & $\beta$ & $\tau$ & $s, m$ & $p, q$ & $\beta, N$ & $\eta, \tau$ \\
    \midrule
    \multirow{2}{*}{Breakfast}  & MSTCN & 0.5 & 0.9 & 0.5 & 1.0, \ 0.5 & 0.4, \ 0.5  & 8, \ 3 & 0.5, \ 0.5 \\
      & ASFormer & 1.5 & 0.9 & 0.1 & 1.0, \ 1.5 & 0.1, \ 0.5 & 8, \ 3 & 0.1, \ 0.3 \\ \hline
    \multirow{2}{*}{YouTube}  & MSTCN & 1.0 & 0.9 & 0.3 & 1.0, \ 0.5 & 0.1, \ 0.5  & 4, \ 3 & 0.1, \ 0.5 \\
      & ASFormer & 1.0 & 0.9 & 0.3 & 1.0, \ 1.5 & 0.1, \ 1.5 & 8, \ 3 & 0.1, \ 0.3 \\ \hline
      \multirow{2}{*}{GTEA}  & MSTCN & - & 0.999 & 0.7 & - & -  & - & 1.0, \ 0.5 \\
      & ASFormer & - & 0.9 & 0.5 & - & - & - & 1.0, \ 0.1 \\ \hline
      \multirow{2}{*}{50salads}  & MSTCN & - & 0.9 & 0.5 & -& -  & - & -, \ 0.3 \\
      & ASFormer & - & 0.99 & 0.7 & - & - & - & -, \ 0.3 \\
      \hline
      \multirow{2}{*}{Assembly101}  & MSTCN & - & 0.99 & 0.1 & - & -  & - & 1.0, \ 0.1 \\
      & ASFormer & - & 0.9 & 0.3 & - & - & - & 1.0, \ 0.3 \\
    \bottomrule
  \end{tabular}}
\end{table*}

\begin{table}[htb]
\caption{Comparisons on YouTube with harmonic mean of head and tail classes over 3 runs.}
\centering 
\setlength{\tabcolsep}{2mm}{
\resizebox{0.98\columnwidth}{!}{
\begin{tabular}{l|lll|lll}
\hline
\multirow{2}{*}{\textbf{Model}} & \multicolumn{3}{c|}{\textbf{Frame acc}} & \multicolumn{3}{c}{\textbf{Segment F1@25}} \\
\cline{2-7} & \textbf{Head} & \textbf{Tail} & \textbf{Hmean} & \textbf{Head} & \textbf{Tail} & \textbf{Hmean} \\
 \hline
 \textbf{AsFormer} &  53.1 & 17.2 & 26.0 & 47.6 & 20.2 & 28.4 \\
 + G-TLA(ours) &  \textbf{55.4}\tiny{$\pm$0.8} & \textbf{24.0}\tiny{$\pm$0.6} & \textbf{33.5}\tiny{$\pm$0.5} & \textbf{47.3}\tiny{$\pm$0.6} & \textbf{25.3}\tiny{$\pm$0.7} & \textbf{33.0}\tiny{$\pm$0.6} \\
 \hline
 \textbf{MSTCN} & 46.0 & 15.5 & 23.2 & 39.0 & 16.8 & 23.5 \\
 + G-TLA(ours) & \textbf{48.7}\tiny{$\pm$1.0} & \textbf{21.8}\tiny{$\pm$0.8} & \textbf{30.0}\tiny{$\pm$0.8} & \textbf{41.7}\tiny{$\pm$0.4} & \textbf{20.1}\tiny{$\pm$0.5} & \textbf{27.1}\tiny{$\pm$0.5} \\ 
\hline
\end{tabular}
}} 
\label{tab:summary2_old_supp}
\end{table}

\begin{table}[htb]
\caption{Comparison %of different long-tail methods 
on Breakfast with harmonic mean on the head and tail classes over 3 runs.}
\centering 
\setlength{\tabcolsep}{2mm}{
\resizebox{0.98\columnwidth}{!}{
\begin{tabular}{l|lll|lll}
\hline
\multirow{2}{*}{\textbf{Model}} & \multicolumn{3}{c|}{\textbf{Frame acc}} & \multicolumn{3}{c}{\textbf{Segment F1@25}} \\
\cline{2-7} & \textbf{Head} & \textbf{Tail} & \textbf{Hmean} & \textbf{Head} & \textbf{Tail} & \textbf{Hmean} \\
 \hline
 \textbf{AsFormer} & 69.7 & 39.8 & 50.7 & 69.9 & 43.9 & 53.9 \\
 + G-TLA(ours) &  \textbf{70.3}\tiny{$\pm$0.1} & \textbf{43.2}\tiny{$\pm$0.5} & \textbf{53.3}\tiny{$\pm$0.5} & \textbf{71.7}\tiny{$\pm$0.2} & \textbf{46.5}\tiny{$\pm$0.1} & \textbf{56.5}\tiny{$\pm$0.1} \\
 \hline
 \textbf{MSTCN} &  65.1 & 37.7 & 47.7 & 53.3 & 38.7 & 44.8 \\
 + G-TLA(ours) & \textbf{67.6}\tiny{$\pm$0.4} & \textbf{43.0}\tiny{$\pm$0.6} & \textbf{52.7}\tiny{$\pm$0.6} & \textbf{60.1}\tiny{$\pm$0.8} & \textbf{45.2}\tiny{$\pm$0.4} & \textbf{51.5}\tiny{$\pm$0.6} \\
\hline
\end{tabular}}}
\label{tab:summary1_old_supp}
\end{table}

\begin{table}[H]
\caption{Global and balanced result summary for Breakfast.}
\centering
\begin{tabular}{c|ccccc|cccc}
\hline
\multirow{2}{*}{\textbf{Model}} & \multicolumn{5}{c|}{\textbf{Global}} & \multicolumn{4}{c}{\textbf{Balanced}}    \\ 
\cline{2-10} & \textbf{  Edit  } & \textbf{  Acc  } & \textbf{F1@10} & \textbf{F1@25} & \textbf{F1@50} & \textbf{  Acc  } & \textbf{F1@10} & \textbf{F1@25} & \textbf{F1@50} \\ 
\hline
\textbf{MSTCN}                           & 66.6 & 67.7 & 63.2 & 57.9 & 46.0 & 47.7 & 48.3 & 44.8 & 36.9 \\
+ CB~\cite{cui2019class}                 & 66.8 & 67.4 & 63.6 & 57.9 & 45.7 & 48.8 & 49.2 & 45.6 & 37.3 \\
+ Focal~\cite{lin2017focal}              & 67.3 & 68.5 & 63.1 & 57.5 & 45.5 & 46.7 & 48.4 & 44.4 & 35.6 \\
+ BAGS~\cite{li2020overcoming}           & 66.3 & 68.5 & 65.1 & 59.8 & 47.5 & 49.4 & 51.1 & 47.4 & 38.5 \\
+ $\tau$-norm~\cite{kang2019decoupling}  & 66.3 & 67.9 & 62.4 & 57.0 & 45.1 & 46.6 & 47.1 & 43.8 & 35.8 \\
+ LA~\cite{menon2020long}                & 67.2 & 67.6 & 63.1 & 57.9 & 45.6 & 49.8 & 49.0 & 45.7 & 36.8 \\
+ LDAM~\cite{cao2019learning}            & 67.1 & 67.5 & 63.4 & 58.1 & 46.1 & 48.0 & 49.1 & 45.6 & 37.4 \\
+ Seesaw~\cite{wang2021seesaw}           & 67.4 & 68.6 & 63.1 & 57.8 & 46.2 & 50.1 & 48.8 & 45.3 & 37.2 \\
+ G-TLA(ours)                            & \textbf{71.3} & \textbf{70.3} & \textbf{68.3} & \textbf{62.9} & \textbf{50.0} & \textbf{52.7} & \textbf{54.5} & \textbf{51.5} & \textbf{41.5} \\
\hline
\textbf{ASFormer}                         & 74.5 & 72.4 & 75.5 & 69.9 & 56.1 & 50.7 & 57.1 & 53.9 & 44.6 \\
+ CB~\cite{cui2019class}                  & 74.9 & 71.9 & 75.6 & 69.7 & 55.8 & 51.6 & 57.9 & 54.9 & 45.6 \\
+ Focal~\cite{lin2017focal}               & 75.4 & 72.3 & 76.1 & 70.4 & 56.2 & 50.2 & 58.1 & 55.1 & 44.9 \\
+ BAGS~\cite{li2020overcoming}            & 73.7 & 71.8 & 74.7 & 68.9 & 55.9 & 51.2 & 58.0 & 54.9 & 45.7 \\
+ $\tau$-norm~\cite{kang2019decoupling}   & 73.6 & 72.2 & 74.9 & 69.1 & 55.7 & 51.5 & 57.1 & 54.2 & 45.2 \\
+ LA~\cite{menon2020long}                 & 74.9 & 72.5 & 75.6 & 69.7 & 56.3 & 51.3 & 58.6 & 54.9 & 46.0   \\
+ LDAM~\cite{cao2019learning}            & 75.3 & \textbf{72.6} & 76.0 & 70.6 & \textbf{57.2} & 51.7 & 58.2 & 55.1 & 46.6 \\
+ Seesaw~\cite{wang2021seesaw}           & 74.9 & 72.5 & 75.7 & 70.1 & 56.3 & 51.8 & 58.6 & 55.5 & 45.8 \\
+ G-TLA(ours)                            & \textbf{75.5} & 72.2 & \textbf{76.2} & \textbf{70.9} & 56.8 & \textbf{53.3} & \textbf{59.2} & \textbf{56.5} & \textbf{47.5} \\
\hline
\end{tabular}
\label{tab:new_metric_breakfast_supp}
\end{table}

\subsection*{E. \quad Additional Results}
\label{sec: E}

\subsubsection{Main results.} We provide benchmark results with standard deviation over 3 runs in \cref{tab:summary1_old_supp} and \cref{tab:summary2_old_supp}. Each run is over 4 or 5 splits(depending on the dataset). We also show the detailed results on global metrics(Acc, F1@10, F1@25, F1@50, and edit score) and balanced metrics(Acc, F1@10, F1@25, F1@50) in \cref{tab:new_metric_breakfast_supp} and \cref{tab:new_metric_youtube_supp}.  Our method consistently outperforms others across datasets and backbones. These results affirm the effectiveness of our approach in mitigating over-segmentation while concurrently improving balanced metrics. The additional results on F1 scores at IoU thresholds of 0.10 and 0.50 further reveal the consistent trend of F1 scores at different thresholds across various methods.

\begin{table}[htb]
\caption{Global and balanced result summary for Youtube.}
\centering
\begin{tabular}{c|ccccc|cccc}
\hline
\multirow{2}{*}{\textbf{Model}} & \multicolumn{5}{c|}{\textbf{Global}} & \multicolumn{4}{c}{\textbf{Balanced}}    \\ 
\cline{2-10} & \textbf{  Edit  } & \textbf{  Acc  } & \textbf{F1@10} & \textbf{F1@25} & \textbf{F1@50} & \textbf{  Acc  } & \textbf{F1@10} & \textbf{F1@25} & \textbf{F1@50} \\ 
\hline
\textbf{MSTCN}                           & 51.9 & \textbf{68.0} & 44.7 & 39.1 & 23.7 & 23.2 & 27.5 & 23.5 & 15.0 \\
+ CB~\cite{cui2019class}                   & 50.9 & 66.3 & 44.9 & 38.7 & 24.0 & 25.9 & 27.7 & 23.3 & 15.7 \\
+ Focal~\cite{lin2017focal}                & \textbf{52.1} & 67.5 & 45.4 & 40.0 & 24.3 & 25.0 & 28.6 & 25.4 & 15.5 \\
+ BAGS~\cite{li2020overcoming}             & 50.8 & 67.2 & 45.7 & 40.1 & \textbf{25.4} & 25.3 & 28.6 & 24.4 & 15.8 \\
+ $\tau$-norm~\cite{kang2019decoupling}    & 50.0 & 67.3 & 43.8 & 38.0 & 23.6 & 24.5 & 26.7 & 22.9 & 15.4 \\
+ LA~\cite{menon2020long}                  & 49.6 & 67.0 & 44.4 & 38.8 & 23.2 & 25.5 & 27.2 & 22.8 & 14.4 \\
+ LDAM~\cite{cao2019learning}             & 51.3 & 67.6 & 45.2 & 39.2 & 23.6 & 23.2 & 28.0 & 23.7 & 13.7 \\
+ Seesaw~\cite{wang2021seesaw}            & 51.5 & 67.9 & 45.4 & 39.7 & 23.2 & 25.0 & 27.6 & 24.1 & 14.0 \\
+ G-TLA(ours)                             & 51.7 & 67.6 & \textbf{45.8} & \textbf{40.2} & 24.9 & \textbf{30.0} & \textbf{30.9} & \textbf{27.1} & \textbf{17.2} \\
\hline
\textbf{ASFormer}                           & \textbf{59.8} & 69.8 & 51.8 & 45.6 & 29.1 & 26.0 & 32.7 & 28.4 & 19.1 \\
+ CB~\cite{cui2019class}                   & 57.7 & 69.6 & 51.3 & 45.4 & 28.7 & 28.8 & 33.7 & 29.2 & 19.4 \\
+ Focal~\cite{lin2017focal}                & 59.3 & 69.7 & 52.5 & \textbf{46.6} & 29.8 & 26.1 & 35.2 & 29.8 & 18.2 \\
+ BAGS~\cite{li2020overcoming}             & 57.0 & 69.3 & 51.1 & 45.1 & 29.6 & 29.3 & 34.1 & 30.0 & 20.4 \\
+ $\tau$-norm~\cite{kang2019decoupling}    & 58.4 & 69.0 & 50.8 & 44.3 & 28.7 & 27.6 & 32.2 & 28.5 & 19.0 \\
+ LA~\cite{menon2020long}                  & 56.3 & 67.9 & 51.3 & 45.1 & 27.5 & 31.3 & 35.0 & 30.5 & 19.7 \\
+ LDAM~\cite{cao2019learning}             & 57.8 & 69.0 & 51.1 & 44.8 & 28.4 & 29.4 & 33.8 & 29.3 & 19.2 \\
+ Seesaw~\cite{wang2021seesaw}            & 58.5 & 69.2 & 52.0 & 45.7 & 29.4 & 28.0 & 33.5 & 29.2 & 18.1 \\
+ G-TLA(ours)                             & 58.9 & \textbf{69.9} & \textbf{52.8} & 46.2 & \textbf{29.9} & \textbf{33.5} & \textbf{38.8} & \textbf{33.0} & \textbf{22.5} \\
\hline
\end{tabular}
\label{tab:new_metric_youtube_supp}
\end{table}

\subsubsection{Extra plots for YouTube. }
\begin{figure}[htb]
 \centering
 \subfloat[][MSTCN.]{\includegraphics[scale=0.20]{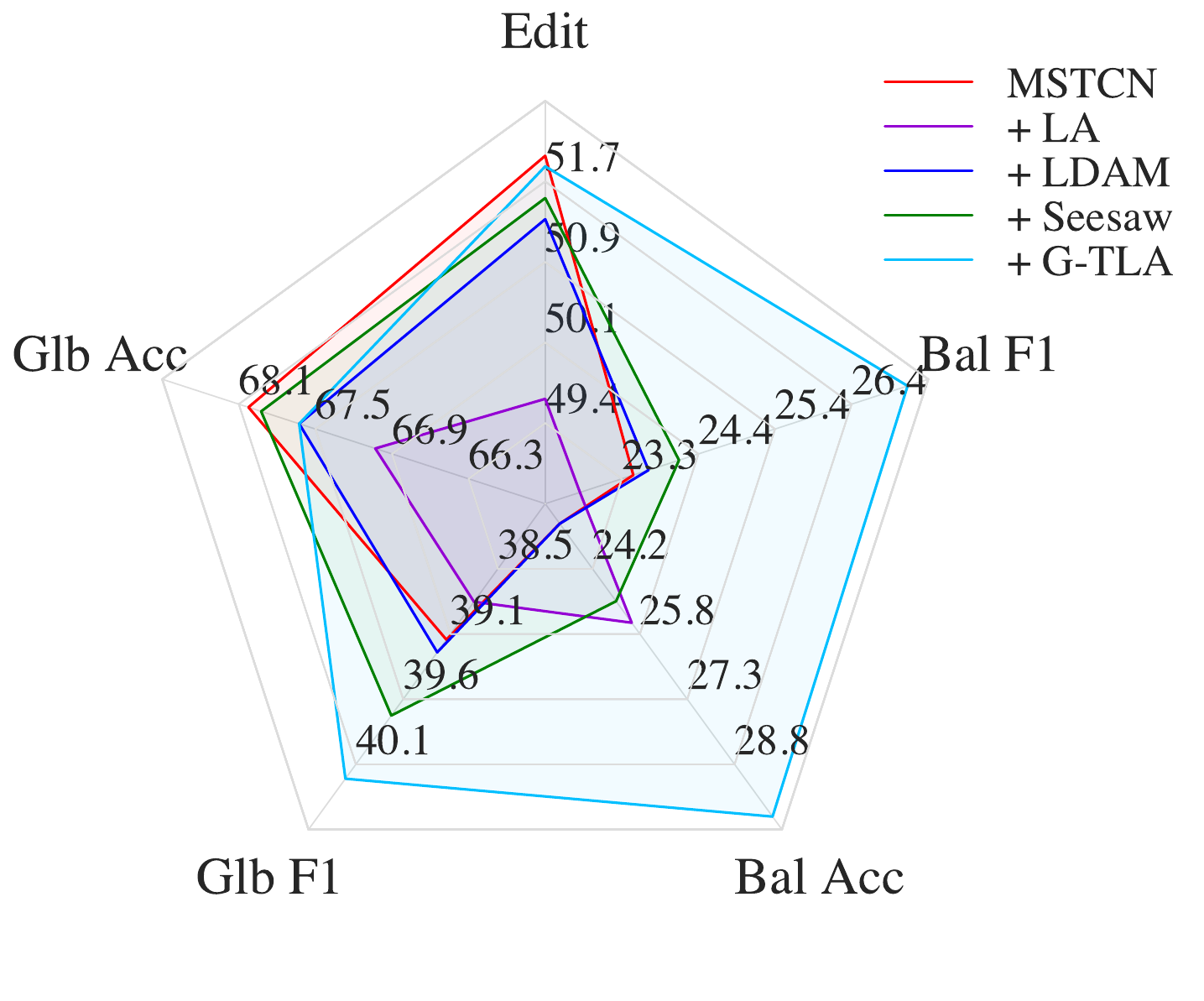}}
 \quad
 \subfloat[][AsFormer.]{\includegraphics[scale=0.20]{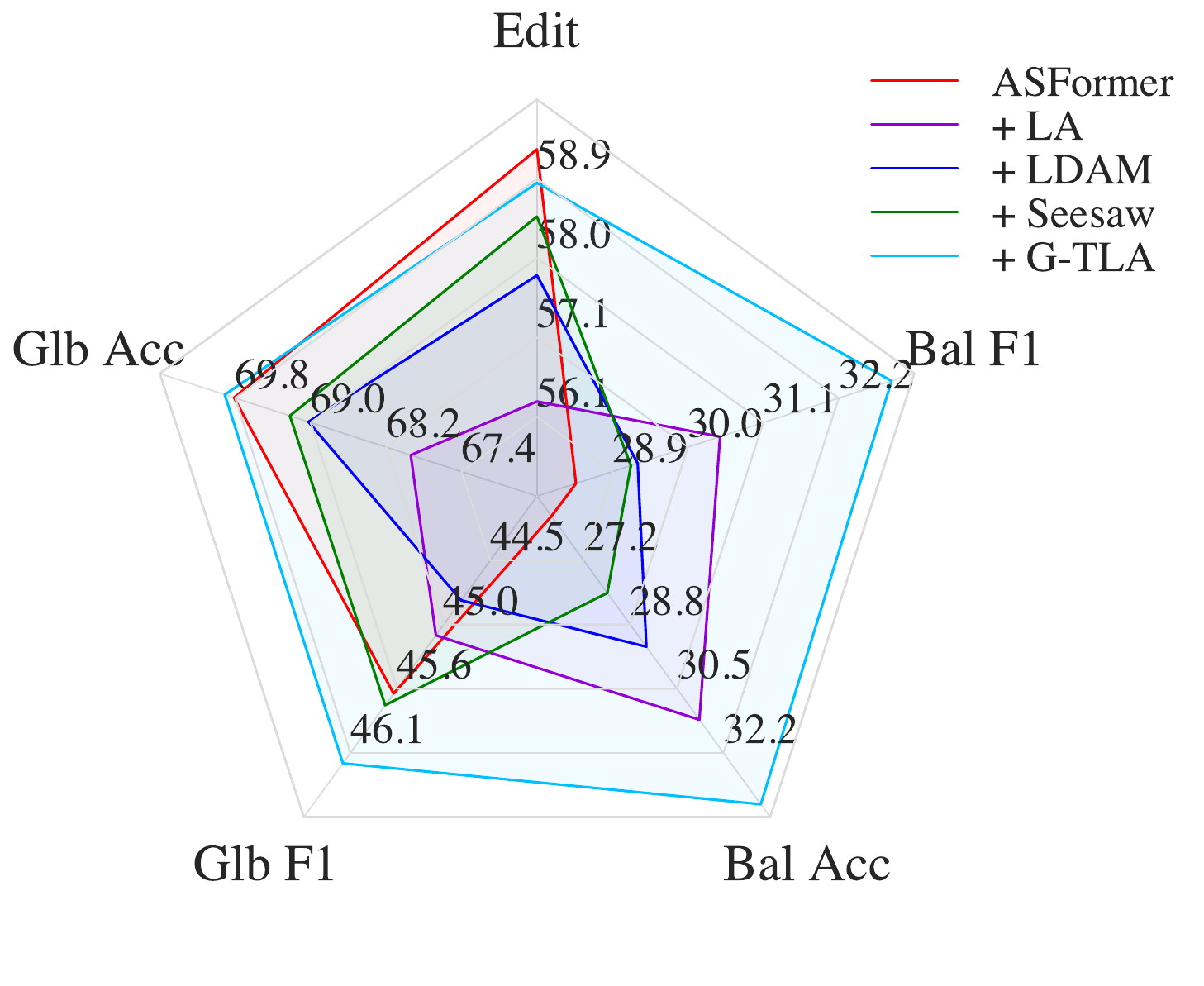}} 
 \caption{Radar charts of logit adjustment methods, measuring the performance along balanced and global metrics on YouTube with MSTCN and AsFormer.}
 \label{fig:radar_mstcn_suppl}
\end{figure}

We further visualize the global and per class results using radar charts on YouTube dataset in \cref{fig:radar_mstcn_suppl}, specifically comparing the performance of logit adjustment methods. The plot demonstrates the superior performance of our method, indicated by the largest enclosed area. Our method excels in segment-wise performance, including edit score and global \& balanced F1 score.

\begin{figure}[htb]
 \centering
 \subfloat[][MSTCN.]{\includegraphics[scale=0.58]{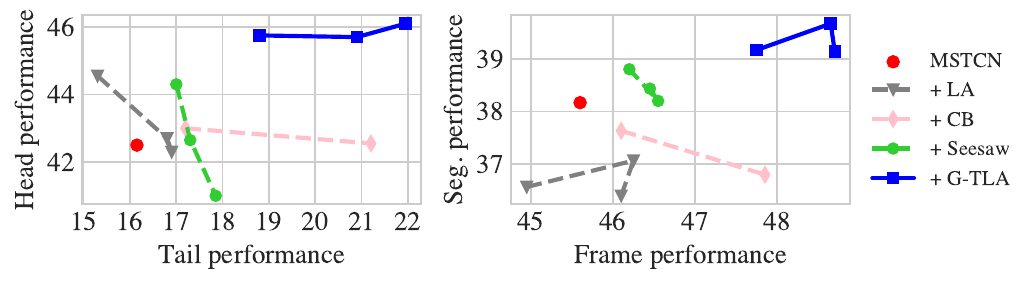}}
 \quad
 \subfloat[][AsFormer.]{\includegraphics[scale=0.58]{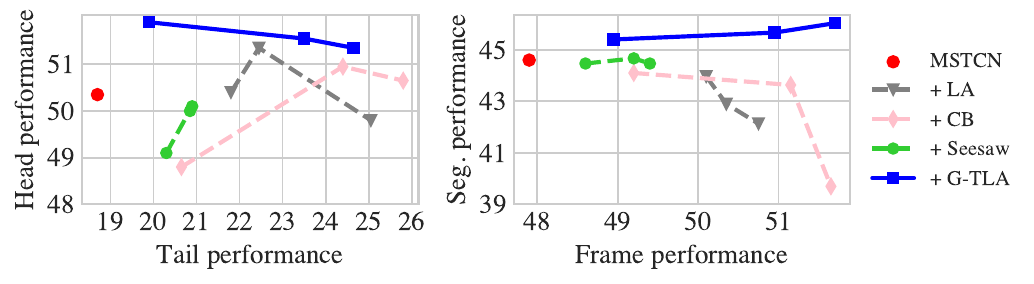}} 
 \caption{Head-Tail \& Frame-Segment trade-offs on YouTube with MSTCN and AsFormer.}
 \label{fig:trade-off_youtube_suppl}
\end{figure}

Long-tail methods for temporal action segmentation exhibit two primary trade-offs: the head-tail trade-off, negatively impacting the head when improving the tail, and the frame-segment trade-off, wherein enhancing tail might adversely affects segment-wise performance. These trade-offs is directly influenced by the hyperparameters. We show these two trade-offs for various methods across backbones in \cref{fig:trade-off_youtube_suppl}.  We fix $\eta = 0.1$ and change $\tau \in \{0.1, 0.3, 0.5\}$ for the trend plotting of our method. For CB and LA, $\beta \in \{0.9, 0.99, 0.999\}$ and $\tau \in \{0.3, 0.5, 0.7\}$. For Seesaw, $q$ is fixed as 0.5, and $p \in \{0.1, 0.2, 0.3\}$. Notably, the curve of our method consistently outperforms others , emphasizing its effectiveness in balancing the learning between head and tail, as well as mitigating over-segmentation.

%%%%%%%%%%%%% ablation study
\subsubsection{Ablation study.}
We present additional ablation studies on YouTube, highlighting the contributions of each component of our G-TLA in \cref{tab:ab_comb_suppl}.  Compared to groupwise classification, temporal logit adjustment yields greater improvement, emphasizing the importance of incorporating temporal priors for YouTube.

The effect of the hyperparameter $\eta$ on YouTube with MSTCN and AsFormer is shown in \cref{tab:eta_supp1} and \cref{tab:eta_supp2}. Small $\eta$ reduces suppression of tail classes, but if too small, it harms group identification during inference. Conversely, a large $\eta$ over-emphasizes the ‘others’ class, harming tail performance. 

\cref{tab:tau_supp1} and \cref{tab:tau_supp2} present the effect of the hyperparameter $\tau$ on YouTube. A smaller $\tau$ represents minimal adjustment, resulting in less improvement for tail classes. Conversely,
a large value of $\tau$ biases towards tail classes and introduces more false positives, causing more over-segmentation.

\begin{table}[htb]
\caption{Ablate group classification(GP), logit adjustment(LA), temporal factor(TF) on YouTube.} 
\centering 
\setlength{\tabcolsep}{1.5mm}{
\resizebox{0.98\columnwidth}{!}{
\begin{tabular}{ccc|ccc|ccc||ccc|ccc}
\hline
\multirow{3}{*}{\textbf{GP}} & \multirow{3}{*}{\textbf{LA}} & \multirow{3}{*}{\textbf{TF}} & \multicolumn{6}{c||}{\textbf{MSTCN}} & \multicolumn{6}{c}{\textbf{ASFormer}} \\ 
\cline{4-15} & & & \multicolumn{3}{c|}{\textbf{Frame acc}} & \multicolumn{3}{c||}{\textbf{Seg. F1}} & \multicolumn{3}{c|}{\textbf{Frame acc}} & \multicolumn{3}{c}{\textbf{Seg. F1}}\\
\cline{4-15} & & & Head & Tail & Hmean & Head & Tail & Hmean & Head & Tail & Hmean & Head & Tail & Hmean\\
\hline
 \xmark & \xmark & \xmark & 46.0 & 15.5 & 23.2 & 39.0 & 16.8 & 23.5 & 53.1 & 17.2 & 26.0 & \textbf{47.6} & 20.2 & 28.4 \\ 
 \xmark & \cmark & \xmark & 46.0 & 17.6 & 25.5 & 39.4 & 16.0 & 22.8 & 53.9 & 22.1 & 31.3 & 46.9 & 22.5 & 30.5 \\
 \cmark & \xmark & \xmark & 46.5 & 16.1 & 23.9 & \textbf{42.5} & 16.8 & 24.0 & 53.1 & 17.6 & 26.4 & 46.9 & 21.8 & 29.7 \\ 
 \cmark & \cmark & \xmark & \textbf{48.9} & 19.9 & 28.3 & 42.3 & 17.7 & 25.0 & 55.3 & 20.6 & 30.0 & 47.1 & 21.8 & 29.8 \\
 \cmark & \cmark & \cmark & 48.7 & \textbf{21.8} & \textbf{30.0} & 41.7 & \textbf{20.1} & \textbf{27.1} & \textbf{55.4} & \textbf{24.0} & \textbf{33.5} & 47.3 & \textbf{25.3} & \textbf{33.0} \\
\hline
\end{tabular}}
}
\label{tab:ab_comb_suppl}
\end{table}

%%%
\begin{table}[t]
 \begin{minipage}{.48\linewidth}
 \caption{Varying $\eta$ for group-wise classification, with fixed number of groups $n=5$ and $\tau=0.5$ on YouTube with MSTCN.}
 \label{tab:eta_supp1}
\centering 
\setlength{\tabcolsep}{1.1mm}{
\resizebox{1\columnwidth}{!}{
 \begin{tabular}{c|ccc|ccc}
 \hline
 \multirow{2}{*}{$\eta$} & \multicolumn{3}{c|}{\textbf{Frame acc}} & \multicolumn{3}{c}{\textbf{Seg. F1}} \\ 
 \cline{2-7} & Head & Tail & Hmean & Head & Tail & Hmean \\ \hline
 0.1 & \textbf{48.7} & \textbf{21.8} & \textbf{30.0} & 41.7 & \textbf{20.1} & \textbf{27.1} \\ 
 0.3 & 48.4 & 18.5 & 26.7 & 42.2 & 16.4 & 23.6 \\ 
 0.5 & 48.4 & 18.3 & 26.5 & \textbf{43.0} & 16.9 & 24.2 \\ 
 \hline
 \end{tabular}}}
 \end{minipage}%
 \quad
 \begin{minipage}{.48\linewidth}
 \centering
 \caption{Varying $\eta$ for group-wise classification, with fixed number of groups $n=5$ and $\tau=0.3$ on YouTube with AsFormer.}
 \label{tab:eta_supp2}
\centering 
\setlength{\tabcolsep}{1.1mm}{
\resizebox{1\columnwidth}{!}{
 \begin{tabular}{c|ccc|ccc}
 \hline
 \multirow{2}{*}{$\tau$} & \multicolumn{3}{c|}{\textbf{Frame acc}} & \multicolumn{3}{c}{\textbf{Seg. F1}} \\ 
 \cline{2-7} & Head & Tail & Hmean & Head & Tail & Hmean \\ \hline
 0.1 & 55.4 & \textbf{24.0} & \textbf{33.5} & 47.3 & \textbf{25.3} & \textbf{33.0} \\ 
 0.3 & \textbf{55.6} & 21.6 & 31.1 & 48.4 & 21.8 & 30.1 \\ 
 0.5 & 55.4 & 19.3 & 30.0 & \textbf{48.6} & 21.5 & 29.8 \\
 \hline
 \end{tabular}}}
 \end{minipage} 
\end{table}

%%%%% 

\begin{table}[htb]
 \begin{minipage}{.48\linewidth}
 \caption{Varying $\tau$ for temporal logit adjustment, with fixed number of groups $n=5$ and $\eta=0.1$ on Youtube with MSTCN.}
 \label{tab:tau_supp1}
\centering 
\setlength{\tabcolsep}{1.1mm}{
\resizebox{1\columnwidth}{!}{
 \begin{tabular}{c|ccc|ccc}
 \hline
 \multirow{2}{*}{$\eta$} & \multicolumn{3}{c|}{\textbf{Frame acc}} & \multicolumn{3}{c}{\textbf{Seg. F1}} \\ 
 \cline{2-7} & Head & Tail & Hmean & Head & Tail & Hmean \\ \hline
 0.1 & 49.4 & 17.1 & 25.4 & 41.1 & 18.2 & 25.3  \\ 
 0.3 & 48.8 & 18.0 & 26.0 & 41.7 & 18.2 & 25.7 \\ 
 0.5 & 48.7 & 21.8 & 30.0 & 41.7 & \textbf{20.1} & \textbf{27.1} \\ 
 0.7 & \textbf{48.9} & \textbf{22.5} & \textbf{30.8} & \textbf{42.2} & 17.9 & 25.4 \\ 
 \hline
 \end{tabular}}}
 \end{minipage}%
 \quad
 \begin{minipage}{.48\linewidth}
 \centering
 \caption{Varying $\tau$ for temporal logit adjustment, with fixed number of groups $n=5$ and $\eta=0.1$ on Youtube with AsFormer}
 \label{tab:tau_supp2}
\centering 
\setlength{\tabcolsep}{1.1mm}{
\resizebox{1\columnwidth}{!}{
 \begin{tabular}{c|ccc|ccc}
 \hline
 \multirow{2}{*}{$\tau$} & \multicolumn{3}{c|}{\textbf{Frame acc}} & \multicolumn{3}{c}{\textbf{Seg. F1}} \\ 
 \cline{2-7} & Head & Tail & Hmean & Head & Tail & Hmean \\ \hline
 0.1 & 54.3 & 18.9 & 27.9 & \textbf{49.5} & 21.1 & 29.7 \\ 
 0.3 & \textbf{55.4} & \textbf{24.0} & \textbf{33.5} & 47.3 & \textbf{25.3} & \textbf{33.0} \\ 
 0.5 & \textbf{55.4} & 23.4 & 32.9 & 46.7 & 23.6 & 30.6 \\ 
 \hline
 \end{tabular}}}
 \end{minipage} 
\end{table}

%%%%
\subsubsection{Group identification results.} 
 Group identification impacts the final performance. Group identification accuracy is shown in \cref{tab:group_id_supp}. The baseline selects predicted groups based on summed output probabilities within each group. Our groupwise classification improves activity identification, reducing false positives and enhancing both frame and segment-wise performance. For datasets without activity label, we use clustering results as ground truth. Our method improves group accuracy from 62.0\%(baseline) to 83.0\% on Assembly101 using MSTCN.

\begin{table}[t]
\caption{Accuracy of group identification}
\centering
\resizebox{0.55\columnwidth}{!}{
\begin{tabular}{c|cc|cc}
\hline
\multirow{2}{*}{\textbf{Method}} & \multicolumn{2}{c|}{\textbf{Breakfast}} & \multicolumn{2}{c}{\textbf{Youtube}} \\
\cline{2-5} & \textbf{MSTCN} & \textbf{ASFormer} & \textbf{MSTCN} & \textbf{ASFormer} \\
\hline
Baseline & 87.2   &  89.1   &  89.3  & 93.1 \\
G-TLA    & 90.1   &  90.2   &  93.4  &  94.2 \\ 
\hline
\end{tabular}
}
\label{tab:group_id_supp}
\end{table}

%%%%
\subsubsection{Results on other datasets}
In \cref{tab:add_sum_50salads}, \cref{tab:add_sum_gtea}, and \cref{tab:add_sum_assembly}, we present our model's performance on other three datasets for completeness. GTEA~\cite{fathi2011learning} and 50Salads~\cite{stein2013combining} have smaller vocabulary sizes and are less imbalanced. Assembly101~\cite{sener2022} is long-tailed but less explored. We do not emphasize Assembly101 as the head classes are also not well-learned. Enhancing the tail classes is less meaningful when the head classes still perform poorly. We include several methods from each type to show the comparison. Our method demonstrates competitive results, further validating its effectiveness. 

\begin{table}[htb]
\caption{Additional results on 50salads.}
\centering 
\setlength{\tabcolsep}{1.6mm}{
\resizebox{0.9\columnwidth}{!}{
\begin{tabular}{l|ccc|ccc|ccc}
\hline
 \multirow{2}{*}{\textbf{Model}} & \multicolumn{3}{c|}{\textbf{Frame acc}} & \multicolumn{3}{c|}{\textbf{Segment F1@25}} & \multicolumn{3}{c}{\textbf{Global}}  \\
\cline{2-10} & \textbf{Head} & \textbf{Tail} & \textbf{Hmean} & \textbf{Head} & \textbf{Tail} & \textbf{Hmean} & \textbf{Edit} & \textbf{F1@25} & \textbf{Acc} \\
 \hline
 \textbf{AsFormer} & 90.6 & 77.4 & 83.5 & 87.5 & 80.3 & 83.8 & 79.0 & 82.3 & 85.2\\
 + CB~\cite{cui2019class} & 90.9 & 78.1 & 84.0  & 88.4 & 81.4 & 84.8 & 78.7 & 83.2 & 85.8 \\
 + $\tau$-norm~\cite{kang2019decoupling} & 90.4 & 77.5 & 83.5 & 87.7 & 80.2 & 83.8 & 78.9 & 82.2 & 85.2 \\
 + LA~\cite{menon2020long} & 90.2 & 78.3 & 83.8 & \textbf{89.8} & 82.1 & 85.7 & 79.8 & 84.5 & 85.4 \\
 + G-TLA(ours) & \textbf{90.8} & \textbf{79.7} & \textbf{84.9} & 89.4 & \textbf{83.1} & \textbf{86.1} & \textbf{80.7} & \textbf{84.6} & \textbf{86.3}\\
 \hline
 \textbf{MSTCN} & 87.7 & 70.0 & 77.9 & 85.7 & 72.1 & 78.3 & 71.4 & 75.9 & 81.1 \\
  + CB~\cite{cui2019class} & 88.4 & 69.3 & 77.7 & 85.3 & 72.0 & 78.1 & 71.1 & 75.5 & 81.0 \\
  + $\tau$-norm~\cite{kang2019decoupling} & 87.6 & 70.3 & 78.0 & 85.1 & 71.6 & 77.8 & 70.8 & 75.3 & 81.1 \\
  + LA~\cite{menon2020long} & 87.5 & 69.6 & 77.5 & 86.0 & 71.0 & 77.8 & 70.4 & 75.2 & 80.8 \\
 + G-TLA(ours) & \textbf{89.0} & \textbf{71.7} & \textbf{79.4} & \textbf{86.8} & \textbf{73.8} & \textbf{79.8} & \textbf{72.0} & \textbf{77.3} & \textbf{81.9}\\
 \hline
\end{tabular}}}
\label{tab:add_sum_50salads}
\end{table}

\begin{table}[htb]
\caption{Additional results on GTEA.}
\centering 
\setlength{\tabcolsep}{1.6mm}{
\resizebox{0.9\columnwidth}{!}{
\begin{tabular}{l|ccc|ccc|ccc}
\hline
 \multirow{2}{*}{\textbf{Model}} & \multicolumn{3}{c|}{\textbf{Frame acc}} & \multicolumn{3}{c|}{\textbf{Segment F1@25}} & \multicolumn{3}{c}{\textbf{Global}}  \\
\cline{2-10} & \textbf{Head} & \textbf{Tail} & \textbf{Hmean} & \textbf{Head} & \textbf{Tail} & \textbf{Hmean} & \textbf{Edit} & \textbf{F1@25} & \textbf{Acc} \\
 \hline
 \textbf{AsFormer} & \textbf{80.6} & 81.7 & 81.2 & 72.5 & 85.4 & 78.4 & \textbf{88.4} & 89.4 & 81.1 \\
 + CB~\cite{cui2019class}  & 79.5 & 84.0 & 81.6 & 71.4 & 88.3 & 78.9 & 86.7 & 89.0 & 80.8 \\
 + $\tau$-norm~\cite{kang2019decoupling} & 80.5 & 82.1 & 81.3 & \textbf{72.8} & 85.2 & 78.5 & 88.3 & 89.5 & 81.1\\
 + LA~\cite{menon2020long}  & 79.5 & 82.7 & 81.1 & 70.9 & 88.4 & 78.7 & 87.5 & 88.9 & 80.5\\
 + G-TLA(ours)  & 80.2 & \textbf{84.5} & \textbf{82.3} & 72.0 & \textbf{90.4} & \textbf{80.2} & 87.9 & \textbf{89.6} & \textbf{81.2}\\
 \hline
 \textbf{MSTCN}  & \textbf{77.6} & 80.3 & 78.9 & 69.4 & 86.6 & 77.0 & 84.8 & 87.2 & 78.0\\
  + CB~\cite{cui2019class}  & 76.2 & 82.6 & 79.3 & 68.8 & 90.1 & 78.0  & 85.3 & 87.7 & 78.5 \\
  + $\tau$-norm~\cite{kang2019decoupling}  & 77.5 & 80.6 & 79.0 & 69.2 & 86.8 & 77.0 & 84.5 & 87.2 & 78.0 \\
  + LA~\cite{menon2020long}  & 77.0 & 83.0 & 79.8 & \textbf{69.8} & 86.3 & 77.2 & 85.4 & 87.2 & 78.5\\
 + G-TLA(ours)  & 77.5  & \textbf{83.7} & \textbf{80.5} & 69.5 & \textbf{90.3} & \textbf{78.5
 }& \textbf{85.8} & \textbf{87.9} & \textbf{78.6}\\
 \hline
\end{tabular}}}
\label{tab:add_sum_gtea}
\end{table}

\begin{table}[htb]
\caption{Additional results on Assembly101.}
\centering 
\setlength{\tabcolsep}{1.6mm}{
\resizebox{0.9\columnwidth}{!}{
\begin{tabular}{l|ccc|ccc|ccc}
\hline
 \multirow{2}{*}{\textbf{Model}} & \multicolumn{3}{c|}{\textbf{Frame acc}} & \multicolumn{3}{c|}{\textbf{Segment F1@25}} & \multicolumn{3}{c}{\textbf{Global}}  \\
\cline{2-10} & \textbf{Head} & \textbf{Tail} & \textbf{Hmean} & \textbf{Head} & \textbf{Tail} & \textbf{Hmean} & \textbf{Edit} & \textbf{F1@25} & \textbf{Acc} \\
 \hline
 \textbf{AsFormer} & 35.2 & 5.7  & 9.8 & 29.0 & 4.8 & 8.2 & \textbf{31.8} & \textbf{30.4} & 41.1 \\
 + CB~\cite{cui2019class}  & 35.4 & 5.9 & 10.1 & 26.5 & 5.2 & 8.7 & 30.6 & 28.2 & 41.0\\
 + $\tau$-norm~\cite{kang2019decoupling}  & 32.2 & 4.9 & 8.5 & 21.8 & 3.2 & 5.6 & 24.3 & 22.7 & 38.5\\
 + LA~\cite{menon2020long}  & 36.1 & 5.9 & 10.1 & 27.5 & 5.7 & 9.4 & 30.2 & 28.5 & \textbf{41.4} \\
 + G-TLA(ours)  & \textbf{36.8} & \textbf{9.2} & \textbf{14.7} & \textbf{30.7} & \textbf{8.3} & \textbf{13.1} & 30.7 & 29.8 & 41.0\\
 \hline
 \textbf{MSTCN}  & 33.9 & 4.7 & 8.2 & 26.3 & 3.9 & 6.8 & 30.1 & 27.2 & \textbf{39.8} \\
  + CB~\cite{cui2019class}  & 32.5 & 7.3 & 11.9 & 27.1 & 4.5 & 7.7 & 25.1 & 22.4 & 37.9 \\
  + $\tau$-norm~\cite{kang2019decoupling}  & 34.0 & 4.3 & 7.6 & 25.9 & 4.2 & 7.2 & 30.5 & 27.4 & 39.6 \\
  + LA~\cite{menon2020long}  & 34.1 & 7.4 & 12.1 & 27.3 & 5.6 & 9.3 & 30.0 & 26.2 & 39.5 \\
 + G-TLA(ours)  & \textbf{34.9} & \textbf{8.0} & \textbf{13.0} & \textbf{30.2} & \textbf{5.8} & \textbf{9.7} & \textbf{30.5} & \textbf{28.5} & 39.2\\
 \hline
\end{tabular}}}
\label{tab:add_sum_assembly}
\end{table}